\documentclass{article} % For LaTeX2e
\usepackage{iclr2026_conference,times}

\usepackage{amsmath,amsfonts,bm}

\def\eqref#1{equation~\ref{#1}}
\def\1{\bm{1}}

\DeclareMathAlphabet{\mathsfit}{\encodingdefault}{\sfdefault}{m}{sl}
\SetMathAlphabet{\mathsfit}{bold}{\encodingdefault}{\sfdefault}{bx}{n}

\usepackage{hyperref}
\usepackage{url}
\usepackage{float} 
\usepackage{booktabs}
\usepackage{graphicx}
\usepackage{amsmath}
\usepackage{multirow}
\usepackage{threeparttable}
\usepackage{tabularx}
\usepackage{longtable}
\usepackage{stfloats}
\usepackage{cleveref}

\usepackage{marvosym}

\newcommand{\ie}{i.e.}

\usepackage{caption}

\title{LLaVA-RadZ: Can Multimodal Large Language Models Effectively Tackle Zero-shot Radiology Recognition?}

\iclrfinalcopy
\author{Bangyan Li$^1$\thanks{Equal contribution. \qquad\qquad\textsuperscript{\Letter}Corresponding author.}\quad
    Wenxuan Huang$^{1,2}$\footnotemark[1]\, \textsuperscript{\Letter},
    Zhenkun Gao$^1$,
    Yeqiang Wang$^3$, 
    \textbf{Yunhang Shen}$^4$, \\
    \textbf{Jingzhong Lin}$^1$, 
    \textbf{Ling You}$^1$,
    \textbf{Yuxiang Shen}$^5$,
    \textbf{Shaohui Lin}$^1$\textsuperscript{\Letter},
    \textbf{Wanli Ouyang}$^2$,
    \textbf{Yuling Sun}$^{1}$\textsuperscript{\Letter} \\
    {\normalsize$^1$East China Normal University} \qquad
    {\normalsize$^2$The Chinese University of Hong Kong} \\
    {\normalsize$^3$Northwest A\&F University} \qquad
    {\normalsize$^4$Tencent Youtu Lab.} \qquad
    {\normalsize$^5$Xiamen University} \\
    {\tt\small \{libangyan2878, osilly0616, shaohuilin007, yulingsun.lv\}@gmail.com}
}
\begin{document}
\maketitle
\begin{abstract}
%Zero-shot disease recognition has emerged as a cutting-edge research topic in medical artificial intelligence.
%
% To reduce reliance on annotated medical images, existing studies typically employ paired image-report data for contrastive learning, enabling fine-grained cross-modal alignment and achieving significant progress.
% To reduce reliance on expensive manual annotations, 
%Most existing studies typically employ paired image-report samples with contrastive learning to reduce reliance on expensive manual annotations, enabling fine-grained cross-modal alignment and achieving significant progress.
%
Recently, Multimodal Large Language Models (MLLMs) have demonstrated exceptional capabilities in visual understanding and reasoning across various vision-language tasks.
However, we found that MLLMs cannot process effectively from fine-grained medical image data in the traditional Visual Question Answering (VQA) pipeline, as they do not exploit the captured features and available medical knowledge fully, results in MLLMs usually performing poorly in zero-shot medical disease recognition.
Fortunately, this limitation does not indicate that MLLMs are fundamentally incapable of addressing fine-grained recognition tasks.
From a feature representation perspective, MLLMs demonstrate considerable potential for tackling such challenging problems.
Thus, to address this challenge, we propose \textbf{\textit{LLaVA-RadZ}}, a simple yet effective framework for zero-shot medical disease recognition via utilizing the existing MLLM features.
Specifically, we design an end-to-end training strategy, termed \textit{Decoding-Side Feature Alignment Training (\textbf{DFAT})} to take advantage of the characteristics of the MLLM decoder architecture and incorporate modality-specific tokens tailored for different modalities.
Additionally, we introduce a \textit{Domain Knowledge Anchoring Module (\textbf{DKAM})} to exploit the intrinsic medical knowledge of large models, which mitigates the \textit{category semantic gap} in image-text alignment.
Extensive experiments demonstrate that our LLaVA-RadZ significantly outperforms traditional MLLMs in zero-shot disease recognition, achieving the comparable performance to the well-established and highly-optimized CLIP-based approaches.
\end{abstract}

% 医学零样本疾病识别已成为医学人工智能领域的前沿研究课题。为减少对医学图像标注的依赖，现有研究通常采用成对的图像-报告数据进行对比学习，从而实现细粒度的跨模态对齐，并取得了显著进展。近年来，多模态大模型（MLLM）在各类视觉-语言任务中展现出了卓越的视觉理解与推理能力。然而，在医学零样本疾病识别任务中，MLLM的表现仍存在一定局限性。
% 为此，我们提出了一种新颖的医学零样本疾病识别框架——LLaVA-RadZ。具体而言，我们设计了一种端到端训练策略，使模型能够充分利用图像和文本特征，实现更高效的跨模态对齐。此外，我们引入了领域知识锚定模块（Domain Knowledge Anchoring Module, DKAM），该模块利用大模型自身的医学领域知识，有效缓解了图文对齐过程中存在的类别语义鸿沟（category semantic gap），从而提高类别级对齐的准确性，实现更精确的疾病识别。
% 我们在五个医学影像下游数据集上进行了广泛实验。实验结果表明，与传统 MLLM 方法相比，我们的方法在零样本疾病识别任务上表现出显著优势，并在性能上与当前领域的成熟方法具有一定的竞争力。

% 传统sft训练在我们领域这一任务表现不佳，我们提出了一种新的训练方式    
\section{Introduction}
\label{sec:intro}

With the rapid advancement of deep learning technologies, an increasing number of studies have focused on their applications in medical disease diagnosis, yielding remarkable results \citep{chan2020computer,jamshidi2020artificial,lee2022deep,tran2021deep}.
However, these approaches typically rely on high-quality annotations provided by clinical experts.
Unlike natural image datasets, annotating medical images is both costly and time-consuming.
To address this challenge, recent research has explored methods based on paired medical images and textual reports, leveraging contrastive learning techniques.
By minimizing the distance between paired samples while maximizing the distance between unpaired ones, these CLIP-based approaches enable zero-shot disease recognition, thereby reducing reliance on extensive medical data annotation to a certain extent.
In our in-depth investigation of advanced zero-shot disease recognition methods in the medical domain, several representative CLIP-based models~\citep{lai2024carzero,wu2023medklip,zhang2023knowledge,phan2024decomposing} have achieved significant performance improvements leveraging the capabilities of Large Language Models (LLMs) or incorporate expert domain knowledge to some extent, rather than fully leveraging the models' intrinsic understanding capabilities.

Recently, Multimodal Large Language Models (MLLMs)~\citep{achiam2023gpt,team2023gemini,liu2023visual,huang2024dynamic,huang2025vision,you2025timesoccer} have demonstrated remarkable capabilities across various user-oriented vision-language tasks, such as image comprehension and reasoning, offering new possibilities for zero-shot disease recognition in medical applications.
Among these, LLaVA-Med~\citep{li2024llava} has exhibited exceptional domain-specific medical knowledge in dialogue-based tasks, indicating that it possesses a certain degree of medical expertise.
However, a recent study~\citep{zhang2024visually} found that MLLMs, ~\textit{i.e.}, LLaVA~\citep{liu2023visual}) performed significantly worse than CLIP~\citep{radford2021learning} on standard image classification tasks.

\begin{figure*}[t]
    \center{\includegraphics[width=0.95\textwidth]{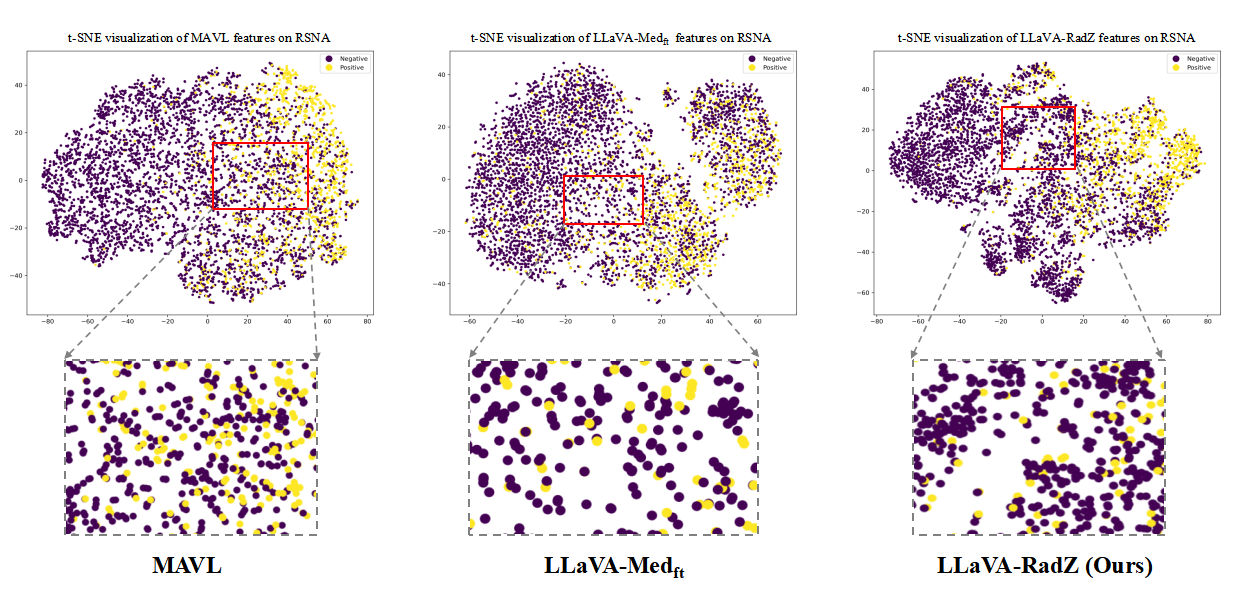}}
    \caption{
        Comparison of Feature Distributions among MAVL, LLaVA-Med$_{\textbf{ft}}$, and LLaVA-RadZ on the RSNA Dataset.
    }
    \label{fig2}
    % \vspace{-8pt}
\end{figure*}

\begin{figure*}[t]
    \center{\includegraphics[width=0.9\textwidth]{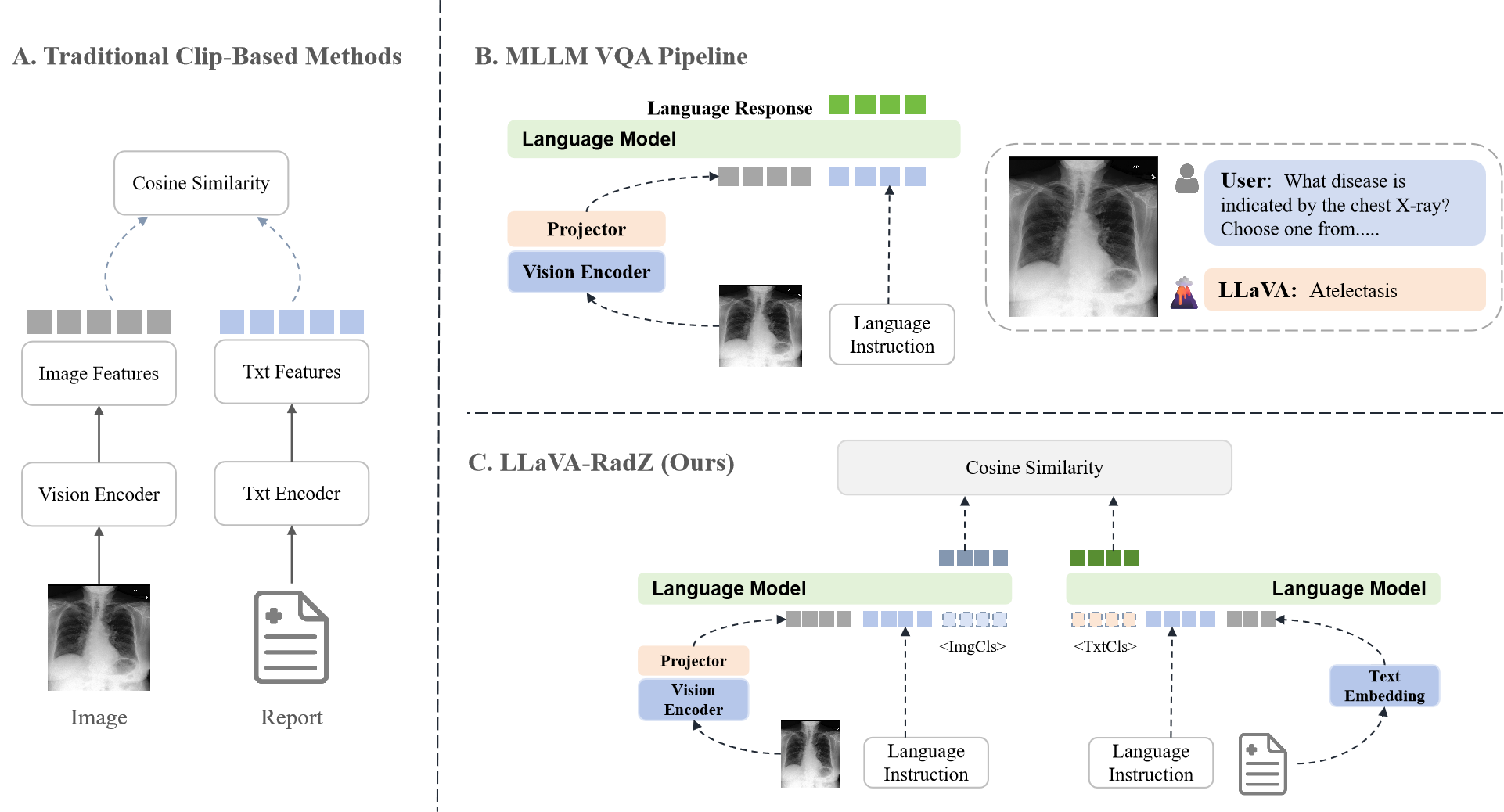}}
    \caption{
        Framework comparison of traditional CLIP-based methods, MLLM VQA pipeline, and the proposed LLaVA-RadZ.
    }
    \label{fig1}
    % \vspace{-8pt}
\end{figure*}

To further validate this observation, we conducted zero-shot classification experiments using multiple MLLMs on five medical imaging datasets (see Tab.~\ref{tab:table1}).
The experimental results are consistent with previous findings, confirming that MLLMs exhibit suboptimal performance in image classification, particularly when dealing with complex medical images.
To enhance the generalization capability of MLLMs in radiology disease recognition tasks, we employed a fine-tuning strategy and performed supervised fine-tuning on the MIMIC-CXR dataset~\citep{johnson2019mimic}.
Additionally, inspired by the work of~\citep{zhang2024visually}, we incorporated a series of optimizations.
While these improvements yielded performance gains, the results remained inferior compared to CLIP-based models.
This phenomenon raises a critical question: \emph{Can MLLMs effectively perform zero-shot disease recognition?}

As shown in Fig.~\ref{fig2}, we visualize the feature distributions of MAVL~\citep{phan2024decomposing}, LLaVA-Med$_{\textbf{ft}}$ (fine-tuned by the same dataset of our LLaVA-RadZ) and LLaVA-RadZ on the RSNA~\citep{shih2019augmenting} dataset.
The results indicate that MLLM exhibits strong feature extraction capabilities, comparable to the well-established MAVL in the domain.
However, in the disease recognition task, MAVL significantly outperforms fine-tuned LLaVA-Med.
We hypothesize that this performance gap arises because MLLMs fail to fully utilize the extracted features for effective disease identification via traditional VQA pipeline.

Inspired by this, we propose a simple yet effective LLaVA-RadZ framework for zero-shot disease recognition using the MLLM features.
Our proposed framework has the fundamental difference compared with previous CLIP-base methods and traditional MLLM VQA pipeline.
As shown in Fig.~\ref{fig1}, we design a dedicated MLLM feature-based framework to address zero-shot medical disease recognition.
Our proposed framework effectively leverages pre-trained MLLM representations to overcome the inherent limitations of the traditional VQA pipeline on this task.
Specifically, firstly, we introduce a new training strategy, Decoding-Side Feature Alignment Training (DFAT).
Specifically, we introduce special tokens for both image and text modalities and leverage the autoregressive generation capability of the decoder architecture to extract global representations of images and texts.
Additionally, we incorporate a cross-modal contrastive loss to optimize the model's ability to learn discriminative features.
Furthermore, to mitigate the semantic category gap encountered during fine-grained alignment between medical images and textual reports, we design a Domain Knowledge Anchoring Module (DKAM).
DKAM utilizes the model's intrinsic medical knowledge to extract the semantic information underlying disease categories, constructing disease description vectors that serve as an intermediary bridge to facilitate the alignment between medical images and textual reports, thereby establishing a stable  relationship.
To further enhance the correlation among medical images, textual reports, and disease categories, a category knowledge-guided loss strengthens the association between similar images and corresponding textual reports.

Our main contributions can be summarized as follows.
\begin{itemize}
    \item 
    We analyze the limitations of current MLLMs in addressing complex fine-grained medical disease recognition tasks, investigate the underlying causes of these constraints, and propose a novel end-to-end feature-based MLLM framework to mitigate these challenges.
    To the best of our knowledge, we are the \textbf{\textit{first}} work in the field of medical disease recognition to explore how to use MLLM features directly to solve complex recognition problems.
    \item 
    We propose the tailored training strategy DFAT, and incorporate a cross-modal contrastive loss to optimize the model's ability to achieve effective alignment between visual and textual features.
    Furthermore, we design a DKAM to leverage MLLM’s intrinsic medical knowledge and effectively mitigate semantic gap in image-text alignment, thereby enhancing category-level alignment.
    \item 
    We conduct extensive experiments on multiple large-scale radiology diagnosis datasets, validating the potential of LLaVA-RadZ in zero-shot disease recognition tasks.
\end{itemize}

\section{Approach}

\subsection{Can Med-LLMs Be Good Medical Classifiers?}
Previous studies have explored the classification capabilities of multimodal large language models (MLLMs), revealing that their performance on image classification tasks is often limited. 
For example, ~\citep{zhang2024visually} investigates the performance differences in classification between MLLMs and CLIP, focusing on factors such as inference strategies, training approaches, and datasets.
Inspired by this work, we extend the exploration to zero-shot tasks in the medical domain.
Unlike natural images and text, the relationship between medical images and reports is more complex.
We seek to investigate whether large medical models, leveraging domain-specific knowledge, can achieve superior performance on medical zero-shot tasks.

We first evaluated two open-source MLLMs, \ie, LLaVA-1.5~\citep{liu2023visual} and LLaVA-Med~\citep{li2024llava}, on five medical datasets in a zero-shot classification setting.
The evaluation followed a general large-model classification approach, where the model selects the correct category from a set of candidate options.
As shown in Tab.~\ref{tab:table1}, these models demonstrated limited performance in disease classification tasks and failed to accurately identify various medical conditions.
Given the potential knowledge limitations of these models, we further assessed the performance of more powerful proprietary MLLMs (~\ie, Qwen2.5-Max~\citep{yang2024qwen2}, Gemini-Pro~\citep{team2023gemini}, and GPT-4o~\citep{achiam2023gpt}) on zero-shot medical disease recognition tasks.
As shown in~\cref{tab:table1}, these models exhibited superior classification capabilities. However, they still lagged behind the state-of-the-art domain-specific methods in medical classification.

To enhance the generalization ability of MLLMs in radiology disease identification, we conducted Supervised Fine-Tuning (SFT) on LLaVA-1.5~\citep{liu2023visual} and LLaVA-Med~\citep{li2024llava} using the publicly available MIMIC-CXR dataset~\citep{johnson2019mimic}.
Surprisingly, the fine-tuned models did not achieve consistent performance improvements across the five datasets.
In some cases, their classification performance even deteriorated.
Further analysis of the model outputs revealed that MLLMs did not always focus on disease-specific information in radiology reports.
Instead, they tended to overlearn the textual structures and linguistic patterns of the reports, which limited their classification capability.
To mitigate this issue, we incorporated the Chain-of-Thought (CoT) prompting strategy and adjusted the model's reasoning approach, inspired by the methodology of~\citep{zhang2024visually}, to optimize the model’s decision-making process.
This approach led to moderate improvements in classification performance on medical datasets.
Although the models have not yet reached optimal performance, the results suggest that MLLMs still hold significant potential for zero-shot medical disease recognition.

% 已有研究对mllm的分类能力进行了探索，发现llm在图像分类任务上性能受限。例如Zhang et al.的研究从推理，训练和数据等方面对llm与clip之间的分类性能差异进行了研究。受益于这篇工作的启发，我们对医学领域的零样本任务进一步进行了探索。不同于自然图像和文本，医学影像和报告之间的关系更加复杂。医疗大模型能受益于其具备的领域知识在医学zero shot任务上达到很好的性能吗？

% 首先，我们选取了两个开源多模态大模型 LLaVA-1.5 和 LLaVA-Med在五个医学数据集上进行了零样本分类评测。评测采用通用的大模型分类方法，即在给定的候选选项中选择正确的类别。实验结果如表 1 所示，这些模型在不同疾病的分类任务上表现有限，未能精准识别各类医学疾病。

% 考虑到模型可能存在的知识局限性，我们进一步评测了更强大的闭源 MLLMs（如 Qwen2.5-Max、Gemini-Pro 和 GPT-4o）在零样本医学疾病识别任务中的表现。正如表 1 所示，这些模型展现出了更优的分类能力，但与当前领域内成熟的专用方法相比仍存在一定差距。

% 为了提升 MLLMs 在放射学疾病识别任务中的泛化能力，我们在 MIMIC-CXR 公开数据集上对 LLaVA-1.5 和 LLaVA-Med 进行了 SFT（Supervised Fine-Tuning）微调。然而，令人意外的是，微调后的模型并未在五个数据集上均获得性能提升，甚至在部分数据集上的分类能力有所下降。进一步分析模型输出后，我们发现 MLLMs 并未始终聚焦于报告中的疾病类别信息，而是一定程度上过度学习了报告的文本结构和语言模式，导致分类能力受限。

% 为缓解这一问题，我们借鉴了 Zhang et al. 的研究方法，引入 Chain-of-Thought（CoT） 推理策略，以优化模型的决策过程，并在一定程度上提升了模型在医学数据集上的分类性能。尽管尚未达到最优水平，但实验结果表明，MLLMs 在零样本医学疾病识别任务中仍具有较强的潜力。

\subsection{Motivation}
As previously discussed, despite possessing a certain level of domain knowledge, medical MLLMs have not yet demonstrated remarkable performance in zero-shot medical tasks.
Even with further instruction tuning, their performance remains inferior to that of existing vision-language models (VLMs).
However, it is noteworthy that modifying the inference strategy leads to significant performance improvements, suggesting that MLLMs are indeed capable of capturing medical image and text features.
Nevertheless, these features have yet to be fully exploited.

To address this limitation, we propose the LLaVA-RadZ framework, introducing a novel end-to-end training strategy, Decoding-Side Feature Alignment Training (DFAT).
This approach leverages the unique properties of the MLLM decoder architecture while incorporating modality-specific special tokens to facilitate effective interaction between medical images and textual features, ultimately achieving more robust cross-modal alignment.
As illustrated in Fig.~\ref{fig2}, we compare the feature distribution of our model with MAVL~\citep{phan2024decomposing}, the current state-of-the-art method, on the RSNA~\citep{shih2019augmenting} dataset.
The results clearly demonstrate that our model achieves better clustering of intra-class samples while enhancing inter-class separation, validating the effectiveness of our approach.
Furthermore, we introduce the Domain Knowledge Anchor Module (DKAM), which harnesses the intrinsic medical knowledge of LLMs to bridge the semantic gap between images and text, enabling more precise disease classification.

% 如前所述，具备一定领域知识的医疗大模型并没有在医学zero shot任务中取得令人震撼的效果。即便经过进一步的指令微调也无法达到与已有vlm模型相当的性能。值得注意的是，在改变推理策略后，模型性能有着显著的提升，这一定程度上反映了MLLM能够较好的捕捉到图像文本特征，但似乎并未充分的利用。对此，我们提出了LLaVA-RadZ框架，设计了新的端到端训练策略，解码端对齐训练，利用MLLM decoder架构特性结合为不同模态设计的特殊标记有效实现医学图像和文本特征的交互，实现更加健壮的跨模态对齐。如图1所示，我们在RSNA数据集上和目前领域sota方法MAVL进行了特征分布比较，可以直观的看出我们的模型使得相同类别的样本更加聚集，不同类别的样本分隔更加明显，验证了我们方法的有效性。此外，我们还引入了Domain Knowledge Anchor Module(DKAM)结构，利用llm的内在医学知识，从而缓解图文对齐中的类别语义差距，实现更精准的疾病识别。

\begin{figure*}[t]
    \center{\includegraphics[width=0.9\textwidth]{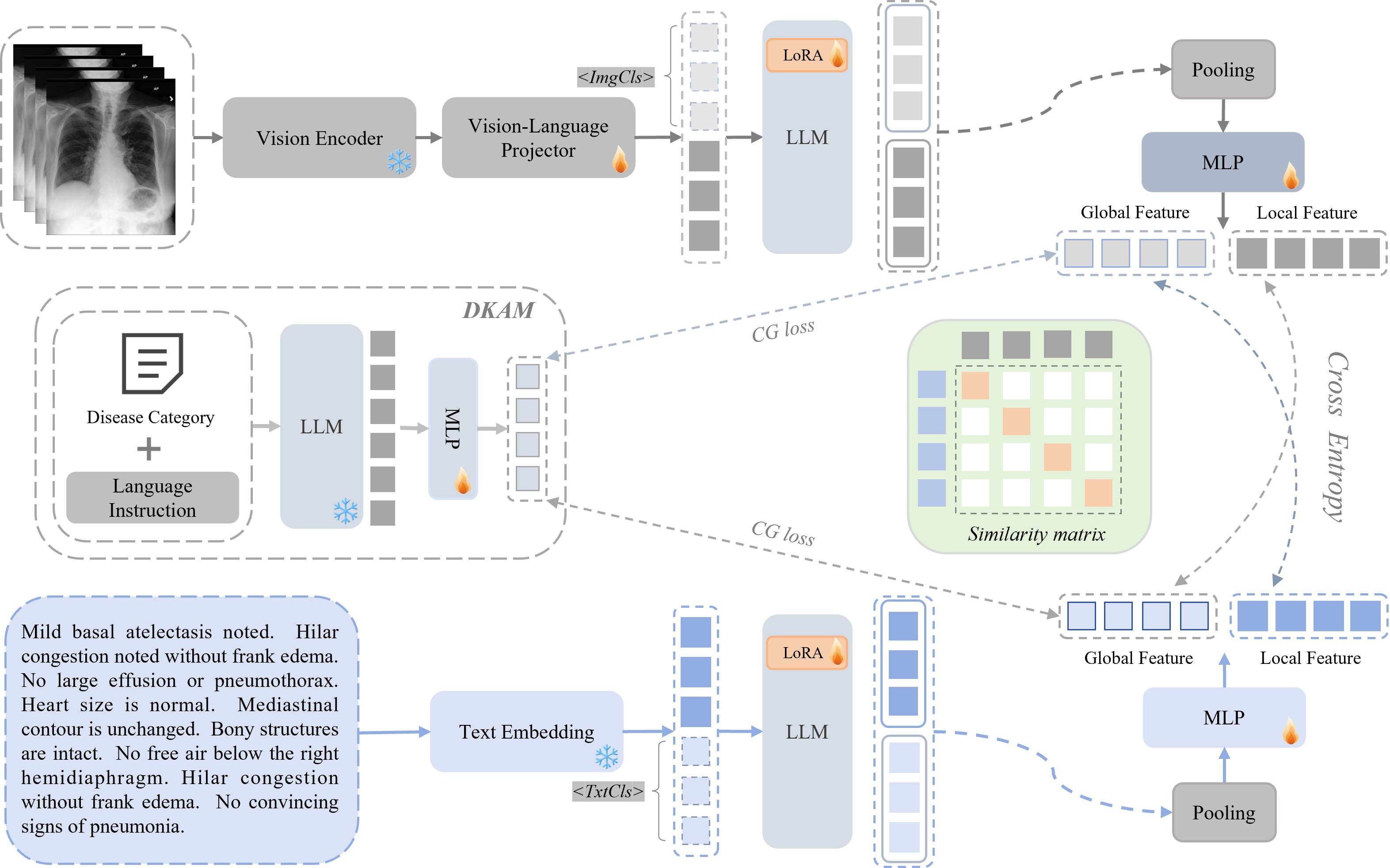}}
    \caption{The LLaVA-RadZ framework consists of three components. (A) Construct a category semantic vector repository using the domain knowledge anchoring module (DKAM). (B) Encode medical images and text, appending $\mathit{\langle ImgCls \rangle}$ and $\mathit{\langle TxtCls \rangle}$ tokens before feeding them into the LLM. (C) Extract global and local features, optimizing with cross-entropy loss, while leveraging the semantic repository for category-level alignment.}
    \label{fig3}
    % \vspace{-8pt}
\end{figure*}

\subsection{The Proposed LLaVA-RadZ}
% \subsubsection{Overview}
We aim to learn generalizable medical image representations from radiology reports to enhance various downstream medical image recognition tasks, particularly when labeled data is scarce.
% Let us assume a training set consisting of $\mathit{N}$ pairs of medical image-text, \ie, $\mathit{S_{\text{train}} = \{(X_\mathrm{1}, Y_\mathrm{1}), \dots, (X_N, Y_N)\}}$, where $\mathit{X}_i\in \mathbb{{R}}^\mathit{H \times W \times 3}$ denotes a medical image, with $\mathit{H}$ and $\mathit{W}$ representing the height and width of the image, respectively. $\mathit{Y}_i$ refers to the corresponding medical text report associated with the image.
The overall framework is illustrated in Fig.~\ref{fig3}.
% Given a pair of medical images $\mathit{X}_i$ and reports $\mathit{Y}_i$, the image and text are first passed through separate visual and text encoders to obtain their respective encoded features.
Given a pair of medical images and reports, the image and text are first passed through separate visual and text encoders to obtain their respective encoded features.
These encoded features and specially designed tokens are then fed into a language model to obtain the final feature representation.
The features are mapped into a common representational space via an MLP projection layer and optimized with the InfoNCE loss.
Furthermore, we propose a Domain Knowledge Anchor Module~(DKAM), which leverages domain knowledge inherent in the model to guide the alignment of text and image features at the category level.

% 我们旨在从放射学报告中学习通用的医学影像表征以惠及各种下游医学影像识别任务，尤其是在标注数据有限的情况下。给定一个包含$\mathit{N}$个医学图像-文本对的训练集,即，其中$\mathit{X_i\in \mathbb{{R}}^\mathit{H \times W \times 3}}$代表一张医学影像，$\mathit{H}$和$\mathit{W}$分别表示图像的高度和宽度。$\mathit{Y_i}$代表医学影像及其对应的医学文本报告。整体框架如图1所示。给定一对医学影像Xi和报告Yi，首先将它们分别输入视觉编码器和文本编码器中获得各自编码后的特征。然后将编码后的特征结合所设计的特殊标记一起输入语言模型中获得最终的特征表示。将特征通过MLP映射到同一个表示空间并引入InfoNCE loss进行优化。此外，我们提出了一个Domain Knowledge Anchor Module，利用模型本身具有的领域知识引导文本特征和图像特征实现类别级对齐。

\subsubsection{End-to-End Training Strategy}
Currently, most MLLMs employ generation-based training objectives for instruction fine-tuning.
Although this approach effectively captures the features of medical images and textual reports, its performance in zero-shot tasks remains suboptimal, as it fails to fully leverage these features.
To address this issue, we propose a novel training strategy, Decoding-Side Feature Alignment Training (DFAT), as illustrated in Fig.~\ref{fig3}.

We consider a training dataset consisting of $\mathit{N}$ pairs of medical image-text samples, denoted as $\mathit{S_{\text{train}} = \{(X_\mathrm{1}, Y_\mathrm{1}), \dots, (X_N, Y_N)\}}$. The medical image 
$\mathit{X}_i\in\mathbb{{R}}^\mathit{H \times W \times 3}$, with $\mathit{H}$ and $\mathit{W}$ representing the height and width of the image, respectively.
$\mathit{Y}_i$ refers to the corresponding medical text report associated with the image.
% Let us consider a training set consisting of $\mathit{N}$ pairs of medical image-text data, \ie, $\mathit{S_{\text{train}} = \{(X_\mathrm{1}, Y_\mathrm{1}), \dots, (X_N, Y_N)\}}$, where $\mathit{X}_i\in\mathbb{{R}}^\mathit{H \times W \times 3}$ denotes a medical image, with $\mathit{H}$ and $\mathit{W}$ representing the height and width of the image, respectively.

% Specifically, we design special tokens for both image and text modalities, where $\mathit{<{ImgCls}_{i}>}(i=0,..,4)$ denotes image feature tokens and $\mathit{<{TxtCls}_{i}>}(i=0,..,8)$ denotes text feature tokens. These special tokens are appended to the image feature extraction instruction $\mathit{{X}_\mathrm{prompt}}$ and the text feature extraction instruction $\mathit{{Y}_\mathrm{prompt}}$, respectively.
Specifically, we design special tokens for both image and text modalities, where $\mathit{<{ImgCls}_{i}>}(i=0,..,4)$ denotes image feature tokens and $\mathit{<{TxtCls}_{i}>}(i=0,..,8)$ denotes text feature tokens.
These special tokens are attached to the image prompt $\mathit{{X}_\mathrm{prompt}}$ and the text prompt $\mathit{{Y}_\mathrm{prompt}}$, respectively.
% The image feature extraction prompt $\mathit{{X}_\mathrm{prompt}}$ follows a format such as 'What disease is indicated by the chest X-ray?', while the text feature extraction prompt $\mathit{{Y}_\mathrm{prompt}}$ follows a format such as 'What disease is described in this text?'. By appending special tokens, we obtain the modified feature extraction prompts $\mathit{{\tilde{X}_\mathrm{prompt}}}$ and $\mathit{{\tilde{Y}_\mathrm{prompt}}}$, which can be represented as:
The image prompt $\mathit{{X}_\mathrm{prompt}}$ has a format similar to ``What disease is indicated by the chest X-ray?'', while the text prompt $\mathit{{Y}_\mathrm{prompt}}$ follows a format such as ``What disease is described in this text?''.
By appending special tokens, we obtain the modified prompts $\mathit{{\tilde{X}_\mathrm{prompt}}}$ and $\mathit{{\tilde{Y}_\mathrm{prompt}}}$, which is represented as:
\begin{equation}
\small
\mathit{{\tilde{X}_\mathrm{prompt}}} = \mathit{{X}_\mathrm{prompt}}+ {\mathit{<{ImgCls}_{i}>}}_{(i=0,..,4)}
,
\label{eq1}
\end{equation}
\begin{equation}
\small
\mathit{{\tilde{Y}_\mathrm{prompt}}} = \mathit{{Y}_\mathrm{prompt}}+ {\mathit{<{TxtCls}_{i}>}}_{(i=0,..,8)}
.
\label{eq2}
\end{equation}
When an image and its corresponding prompt $\mathit{{\tilde{X}_\mathrm{prompt}}}$ are input into the MLLM $\mathcal{F}$ to generate a response $\mathit{{\hat{R}}_\mathrm{img}}$.
Similarly, when a text sample and its corresponding feature extraction prompt $\mathit{{\tilde{Y}_\mathrm{prompt}}}$  are provided as input, the model produces a response $\mathit{{\hat{R}}_\mathrm{txt}}$. This process can be formally expressed as:
\begin{equation}
\small
\begin{aligned}
\mathit{{\hat{R}}_\mathrm{img}^{i}} = \mathcal{F}(\mathit{{X}_{i}},\mathit{{\tilde{X}_\mathrm{prompt}}})
,
\quad 
\mathit{{\hat{R}}_\mathrm{txt}^{i}} = \mathcal{F}(\mathit{{Y}_{i}},\mathit{{\tilde{Y}_\mathrm{prompt}}})
.
\end{aligned}
\label{eq3}
\end{equation}
Due to the autoregressive nature of the decoder architecture, when the LLM processes visual and textual information to generate responses, its internal representations are stored in the designated special tokens.
Specifically, we extract the penultimate layer embedding $\mathit{{\tilde{h}_\mathrm{img}}}$ corresponding to the special token $\mathit{<{ImgCls}_{i}>}$, which stores the global image features ${H}_\mathrm{img}^\mathrm{global}\in \mathbb{R}^{B\times I\times K}$. Here, $\mathit{B}$ denotes the number of image-text pairs in each batch, $\mathit{I}$ represents the number of special image tokens, and $\mathit{K}$ is the dimension of the shared embedding space. After applying a pooling operation followed by an MLP projection layer ${\gamma}_\mathrm{img}$, we obtain the global image feature representation $\mathit{{X}_{g}}\in \mathbb{R}^{B\times K}$. The local image feature $\mathit{{X}_{l}}\in \mathbb{R}^{B\times K}$ is obtained by pooling the hidden states of all tokens except those corresponding to special tokens, followed by an MLP projection layer ${\gamma}_\mathrm{img}$:
\begin{equation}
\small
\begin{aligned}
\mathit{X}_{g} &= {\gamma}_\mathrm{img}(\operatorname{AvgPool}({H}_\mathrm{img}^\mathrm{global}))
, 
\quad
\mathit{X}_{l} = {\gamma}_\mathrm{img}(\operatorname{AvgPool}({H}_\mathrm{img}^\mathrm{local}))
.
\end{aligned}
\label{eq4}
\end{equation}

Similarly, we extract the global text representation $\mathit{{Y}_{g}}\in \mathbb{R}^{B\times K}$ and the local text representation $\mathit{{Y}_{l}}\in \mathbb{R}^{B\times K}$ using the same methodology:
\begin{equation}
\small
\begin{aligned}
\mathit{Y}_{g} &= {\gamma}_\mathrm{txt}(\operatorname{AvgPool}({H}_\mathrm{txt}^\mathrm{global}))
, 
\quad
\mathit{Y}_{l} = {\gamma}_\mathrm{txt}(\operatorname{AvgPool}({H}_\mathrm{txt}^\mathrm{local}))
.
\end{aligned}
\label{eq4}
\end{equation}

To further enhance fine-grained alignment across different modalities, we introduce a cross-modal contrastive loss, $\mathit{{L}_{CA}}$.
Specifically, for the $\mathit{i}$-th image-text pair $(\mathit{{X}_{i}, {Y}_{i}})$ in a batch, we alternately align the global and local features of images and texts.
This procedure yields two symmetric, temperature-normalized InfoNCE objectives: one aligns global image features with local text features, and the other aligns local image features with global text features.
These objectives maximize the mutual information between image-text pairs in the latent space.

For the alignment between global image features and local text features, we calculate two similarity matrices, $\mathit{{S}_{i}^{{X}_{g}\rightarrow {Y}_{l}}}$ and $\mathit{{S}_{i}^{{Y}_{l}\rightarrow {X}_{g}}}$, with the following computation:
\begin{equation}
\small
\begin{aligned}
\mathit{{S}_{i}^{{X}_{g}\rightarrow {Y}_{l}}}=\frac{{X}_{g,i}\cdot {Y}_{l,i}^{T} }{\tau }
,
\quad
\mathit{{S}_{i}^{{Y}_{l}\rightarrow {X}_{g}}}=\frac{{Y}_{l,i}\cdot {X}_{g,i}^{T} }{\tau }
.
\end{aligned}
\label{eq5}
\end{equation}
where $\tau$ is the temperature hyperparameter.
Subsequently, we compute the contrastive loss between the global image and the local text, with the following formula:
\begin{equation}
\small
\begin{aligned}
L_{\text{CA}}^{X_g \rightarrow  Y_l, i}
& = -\log \frac{\exp\bigl(S_i^{X_g \rightarrow  Y_l}\bigr)}
{\sum_{k=1}^{B} \exp\bigl(S_k^{X_g \rightarrow  Y_l}\bigr)}
,
\quad
L_{\text{CA}}^{Y_l \rightarrow  X_g, i} = -\log \frac{\exp\bigl(S_i^{Y_l \rightarrow  X_g}\bigr)}
{\sum_{k=1}^{B} \exp\bigl(S_k^{Y_l \rightarrow  X_g}\bigr)}
.
\end{aligned}
\label{eq5}
\end{equation}
\begin{equation}
\small
\begin{aligned}
L_{\text{CA}}^{X_g \rightarrow  Y_l} = \frac{1}{2}
\sum_{i=1}^{B}
\Bigl(
L_{\text{CA}}^{X_g \rightarrow  Y_l, i}
+ L_{\text{CA}}^{Y_l \rightarrow  X_g, i}
\Bigr)
.
\end{aligned}
\end{equation}
Similarly, for the alignment between local image features and global text features, we compute the contrastive loss between the local image and global text.
\begin{equation}
\small
\begin{aligned}
L_{\text{CA}}^{X_l \rightarrow Y_g}
= -\frac{1}{2}
\sum_{i=1}^{B}
\Bigl(
\log \frac{\exp\bigl(S_i^{X_l \rightarrow Y_g}\bigr)}{\sum_{k=1}^{B} \exp\bigl(S_k^{X_l \rightarrow Y_g}\bigr)} 
+ \log \frac{\exp\bigl(S_i^{Y_g \rightarrow X_l}\bigr)}{\sum_{k=1}^{B} \exp\bigl(S_k^{Y_g \rightarrow X_l}\bigr)}
\Bigr)
.
\end{aligned}
\label{eq5}
\end{equation}
Finally, we obtain our cross-modal contrastive loss $\mathit{{L}_{CA}}$.
\begin{equation}
\small
\begin{aligned}
\mathit{{L}_{CA}}=\frac{1}{2}\left (L_{\text{CA}}^{X_g \rightarrow  Y_l} + L_{\text{CA}}^{X_l \rightarrow Y_g}\right )
.
\end{aligned}
\label{eq5}
\end{equation}

% 目前大多数多模态大语言模型进行指令微调时以生成作为训练目标，这导致模型在进行zero shot任务时性能不够显著，尽管已经很好捕捉到了医学影像和文本报告中的特征却无法充分利用。对此，我们探索了一种新的训练策略——解码端的特征对齐训练,如图1所示。具体来说，我们针对图像和文本两种模态分别设计了特殊标记，其中<ImgClsi>(i=0,..,4)为图像特征标记，<TxtClsi>(i=0,..,8)为文本特征标记。我们将特殊标记分别添加到图像特征提取指令Xprompt和文本特征提取指令Yprompt的末尾。
% Xprompt的格式类似于"What disease is indicated by the chest X-ray?"，文本特征提取指令Yprompt的格式类似于"What disease is described in this text?"。拼接后得到含特殊标记的特征提取指令X'和Y'，可以表示为XXX【列个表达式】。将图像和图像特征提取指令X'共同输入多模态LLM F中，LLM将输出一个响应Rimg,类似的，文本和文本特征提取指令Y'输入到多模态LLM F中,LLM也会输出响应Rtxt，可以表示为Rimg=F(Ximg, X'),Rtxt=F(Ytxt, Y')。

% 由于decoder架构自回归生成特性，当LLM意图解析视觉信息和文本信息生成响应时，其特征会被特殊标记所存储。我们提取<ImgClsi>标记对应的LLM倒数第二层嵌入himg，池化后应用一个MLP投影层rimg获得图像全局特征Xg，将<ImgClsi>标记存储以外的特征池化后经过mlp投影层rimg映射得到图像局部特征Xl,可以表示为:。类似的，我们采用同样的方法获得文本全局特征Yg，文本局部特征Yl，可以表示为:。
% 为了进一步强化不同模态的细粒度对齐能力，我们引入了跨模态对比损失lossca，具体来说，对于batch中的第i个图像-文本对(Xi,Yi)我们交替的将图像和文本的全局以及局部特征进行对齐。这会得到两个对称的温度归一化InfoNCE损失(即全局图像到局部文本对比损失和全局文本到局部图像对比损失)，用于最大化潜在空间中图像和文本对之间的互信息:
% 对于全局图像和局部文本之间的对齐，我们计算得到两个相似度矩阵Si1和Si2，具体计算如下：xxx，进一步计算得到全局图像和局部文本对比损失，计算公式如下:xx,类似的，对于全局文本和局部图像之间的对齐，我们采用相同的方式计算得到全局文本和局部图像对比损失。最终，得到我们的图像和文本跨模态对比损失lossca。
\subsubsection{Domain Knowledge Anchor Module}
In aligning medical images with text reports, we observed that the critical entity of the medical disease categories was merely encoded as features by the model, without considering the underlying semantics.
To address this limitation and further enhance fine-grained alignment capabilities, we introduce the Domain Knowledge Anchoring Module (DKAM).
Initially, we leverage the inherent medical domain expertise of an LLM to generate descriptive explanations for each disease category.
These generated disease descriptions serve as an intermediary bridge to guide the alignment between medical images and text reports.
Specifically, we input the disease list $\mathit{{D}_\mathrm{list}}$ from the training dataset along with a designed prompt template $\mathit{{K}_\mathrm{prompt}}$ into the LLM $\mathcal{F}$.
This process is formally expressed as:
\begin{equation}
\small
\begin{aligned}
\mathit{{\hat{R}}_\mathrm{dis}} = \mathcal{F}(\mathit{{D}_\mathrm{list}},\mathit{{{K}_\mathrm{prompt}}})
.
\end{aligned}
\label{eq3}
\end{equation}

By fully harnessing the LLM's exceptional semantic understanding, we prompt the model to explore the underlying semantics of the disease categories and discern their distinctions, ultimately producing a refined disease description.
The features extracted from the LLM's response are then mapped via a multi-layer perceptron (MLP) to yield the disease description vector $\mathit{\hat{D}}$, which is represented as: 
\begin{equation}
\small
\begin{aligned}
\mathit{\hat{D}} = {\gamma}_\mathrm{dis} \left (\mathit{{\hat{R}}_\mathrm{dis}} \right )
.
\end{aligned}
\label{eq3}
\end{equation}

Subsequently, we introduce the Category of Knowledge-guided Contrastive Loss $\mathit{L_{CG}}$.
Specifically, we calculate the cross-entropy loss between the disease description vector $\mathit{\hat{D}}$ and the global features of both the images $\mathit{X_g}$ and the text $\mathit{Y_g}$. This design encourages the model to better capture the semantic relationships among images, text, and disease categories during training, achieving a more robust category-level alignment.
\begin{equation}
\small
\begin{aligned}
S_i^{\text{img-disease}} = \frac{X_{g,i} \cdot D^T}{\tau}
, 
\quad
S_i^{\text{txt-disease}} = \frac{Y_{g,i} \cdot D^T}{\tau}
.
\end{aligned}
\label{eq3}
\end{equation}
\begin{equation}
\small
\begin{aligned}
L_{\text{CG},i}^{\text{txt}} 
= -\log \frac{\exp\bigl(S_i^{\text{txt-disease}}\bigr)}
{\sum_{k=1}^{N} \exp\bigl(S_k^{\text{txt-disease}}\bigr)}
,
\quad
L_{\text{CG},i}^{\text{img}} 
= -\log \frac{\exp\bigl(S_i^{\text{img-disease}}\bigr)}
{\sum_{k=1}^{N} \exp\bigl(S_k^{\text{img-disease}}\bigr)}
.
\end{aligned}
\label{eq3}
\end{equation}
Here, $\mathit{N}$ represents the number of disease categories, $\mathit{B}$ denotes the number of medical image-text pairs in each batch, and $\tau$ is the temperature hyperparameter. The final category of knowledge-guided loss is as follows:
\begin{equation}
\small
\begin{aligned}
L_{\text{CG}} 
= \frac{1}{2} \sum_{i=1}^{B}
\Bigl(
L_{\text{CG},i}^{\text{txt}} 
+ L_{\text{CG},i}^{\text{img}}
\Bigr)
.
\end{aligned}
\label{eq3}
\end{equation}

By combining the category knowledge-guided loss and the cross-modal contrastive loss, the final objective function is defined as follows:
\begin{equation}
\small
\begin{aligned}
L_{\text{total}} = \lambda L_{\text{CA}} + (1 - \lambda) L_{\text{CG}}
,
\end{aligned}
\label{eq3}
\end{equation}
where $\lambda$ is a balancing factor used to adjust the weights of the two losses, and it is set to $0.5$ by default.
% 在进行医学影像和文本报告的对齐时，我们发现医学疾病类别这一重要实体只是被模型简单的编码为特征，而缺乏了对疾病类别背后语义的深入思考。对此，为了进一步加强细粒度对齐能力，我们引入了领域知识锚点模块即DKAM。首先通过LLM本身具有的医学领域知识对疾病类别进行描述性解释，然后将生成的疾病描述作为中间桥梁引导医学影像和文本报告之间的对齐。具体来说，我们将训练数据集的疾病列表Dlist结合我们所构造的提示模板Kprompt一起输入LLM f中，具体细节见补充材料。通过充分利用LLM在语义理解方面的卓越能力，我们引导模型思考疾病类别背后的语义，以及类别之间的不同点，得到最终的疾病描述。从LLM生成的响应中提取出特征并经过MLP进一步映射成疾病描述向量D，可以表示为xx。接下来我们引入了类别知识引导对比损失LCG。具体来说分别将疾病描述向量D和图像的全局特征Xg以及文本的全局特征Yg进行交叉熵计算，从而使得模型在训练过程中更好的捕捉到图像、文本与疾病类别之间的语义关系，实现类别级对齐。其中N为疾病类别数量，B为每个批次中医学图文对的数量。最终的类别知识引导损失如下：

% 第一部分写探索型方法以及motivation（1页半，
% 我们的不足主要在医疗领域进行了实践，考虑到医疗领域细粒度分类的重要性，暂且没有拓展到其他专有领域。

% 第二部分写自己方法，介绍自己方法，画个框架图，1页半）

% 讨论问题: 大概的故事线，走性能框架方式还是探索性？如果探索性，评估标准是否要修改，需要测多个med-llm？目前已有MLLM分类能力评估采用的大都是ABCD选择形式（目前比较公认的，），但这种效果会很差，不过有论文用这种形式在通用数据集，flower之类的效果测的很高，但这种形式在医学数据集中效果很低。vlmclassifier中几种推理形式都测了，也使用了新的推理策略测试了，类似于clip的策略，但是他在ImageNet的性能仍然显著低于clip，并且说这个实践中开销较大，需要计算每个类别的概率。

% 现有模型 效果不好 ，换了一种推理策略 zero shot分类能力

% 探索性 如何说服使用clip的推理策略更合理，或者也测试ABCD推理策略的性能，但是结构调整了直接load checkpoint测ABCD又不行？

%vlmclassifier论文里提出生成训练目标在学习分类任务时与传统的交叉熵一样有效，这消除了VLM和CLIP之间的性能差距，使得VLM现在成为最先进的分类器（使用的就是sft微调，一个问题，一个答案（label））。可能要argue的点在提升MLLM zero shot性能，即没有明确类别采用传统生成训练目标无法 有效缩短clip与mllm之间的性能差距，分类目标更适合
%
\begin{table*}[t]
\caption{
Comparison of zero-shot disease classification performance of public MLLMs and LLaVA-based exploratory methods across five medical benchmarks.
``ft'' denotes supervised fine-tuning with LoRA, ``CoT'' refers to zero-shot chain-of-thought prompting templates, and ``Inference'' represents CLIP inference strategies. The best results are highlighted in bold and the second-best results are underlined.
}
\label{tab:table1}
\resizebox{\textwidth}{!}{
\begin{tabular}{@{}cl|ccc|ccc|ccc|ccc|ccc@{}}
\toprule
\multicolumn{2}{c|}{\textbf{Dataset}} 
& \multicolumn{3}{c|}{\textbf{CheXpert }} 
& \multicolumn{3}{c|}{\textbf{ChestXray-14}} 
& \multicolumn{3}{c|}{\textbf{COVIDx CXR-2}}  
& \multicolumn{3}{c|}{\textbf{RSNA Pneumonia }} 
& \multicolumn{3}{c}{\textbf{SIIM-ACR }} \\
\midrule 
% \cmidrule(l){1-17} 
\multicolumn{1}{c|}{\textbf{Method}} & \multicolumn{1}{c|}{\textbf{Model}} & AUC $\uparrow$ & F1 $\uparrow$ & ACC $\uparrow$ & AUC $\uparrow$ & F1 $\uparrow$ & ACC $\uparrow$ & AUC $\uparrow$ & F1 $\uparrow$ & ACC $\uparrow$ & AUC $\uparrow$ & F1 $\uparrow$ & ACC $\uparrow$ & AUC $\uparrow$ & F1 $\uparrow$ & ACC $\uparrow$\\ \midrule
\multicolumn{1}{c|}{\multirow{5}{*}{\textbf{MLLM}}} & LLaVA-1.5 (7B)~\citep{liu2023visual} 
&-  &7.50  &8.28  
&-  &3.33  &6.92  
&-  &53.14  &50.28  
&-  &40.53  &55.34  
&-  &23.66  &50.36  \\

\multicolumn{1}{c|}{} & LLavA-Med (7B)~\citep{li2024llava} 
&-  &6.87  &8.94  
&-  &8.02  &6.78  
&-  &34.90  &50.03  
&-  &18.58  &50.00 
&-  &21.91  &49.90  \\

% \multicolumn{1}{c|}{} & LLaVA-OV(8B)~\citep{li2024llavaov} 
% &  &  &  
% &  &  &  
% &  &  &  
% &  &  &  
% &  &  &  \\

\multicolumn{1}{c|}{} & Qwen2.5-Max~\citep{yang2024qwen2} 
&-  &32.23  &67.97  
&-  &19.04  &76.19  
&-  &\underline{75.91}  &\underline{76.81}  
&-  &43.58  &43.59  
&-  &64.70  &\underline{72.57}  \\

% \multicolumn{1}{c|}{} & Claude3~\citep{anthropic2024claude} &  &  &  &  &  &  &  &  &  &  &  &  &  &  &  \\ 
\multicolumn{1}{c|}{} & Gemini-Pro~\citep{team2023gemini} 
&-  &35.01  &76.08  
&-  &14.16  &77.78 
&-  &62.84  &62.90 
&-  &44.23  &51.43  
&-  &61.43  &72.03  \\ 

\multicolumn{1}{c|}{} & GPT-4o~\citep{achiam2023gpt} 
&-  &\underline{45.85}  &\underline{81.14}
&-  &19.85  &\underline{81.55}  
&-  &50.93  &\textbf{77.08}  
&-  &54.20  &65.33  
&-  &64.57  &72.11  \\ 

% \multicolumn{1}{c|}{} & DeepSeek~\citep{guo2025deepseek} &  &  &  &  &  &  &  &  &  &  &  &  &  &  &  \\ 
\midrule
\multicolumn{1}{c|}{\multirow{5}{*}{\textbf{Explorative Methods}}} & LLaVA-1.5-7B$_{\textbf{ft}}$ 
&-  &10.61  &19.62  
&-  &7.85  &19.06  
&-  &27.74  &25.18  
&-  &43.60  &34.80  
&-  &52.37  &50.95  \\

% \multicolumn{1}{c|}{} & LLaVA-1.5-7B$_{\textbf{ft}}$ + CoT &  &  &  &  &  &  &  &  &  &  &  &  &  &  &  \\
% \multicolumn{1}{c|}{} & LLaVA-1.5-7B$_{\textbf{ft}}$ + Linear &  &  &  &  &  &  &  &  &  &  &  &  &  &  &  \\
% \multicolumn{1}{c|}{} & LLaVA-1.5-7B$_{\textbf{ft}}$ + Inference &  &  &  &  &  &  &  &  &  &  &  &  &  &  &  \\
\multicolumn{1}{c|}{} & LLavA-Med-7B$_{\textbf{ft}}$ 
&-  &14.25  &31.46  
&-  &9.00  &21.43 
&-  &27.42  &24.09  
&-  &46.72  &38.88  
&-  &53.11  &57.68  \\

\multicolumn{1}{c|}{} & LLavA-Med-7B$_{\textbf{ft}}$ + CoT~\citep{zhang2024visually}
&-  &8.90  &26.23 
&-  &8.33  &20.46 
&-  &27.12  &26.55  
&-  &49.59  &43.80 
&-  &54.06  &51.07  \\

% \multicolumn{1}{c|}{} & LLavA-Med-7B$_{\textbf{ft}}$ + Linear~\citep{zhang2024visually}
% &  &  &  
% &  &  &  
% &  &  &  
% &  &  &  
% &  &  &  \\

\multicolumn{1}{c|}{} & LLavA-Med-7B$_{\textbf{ft}}$ + Inference~\citep{zhang2024visually}
&71.00  &44.85  &75.45  
&64.30  &\underline{21.73}  &70.86  
&71.07  &69.84  &60.39  
&77.51  &\underline{69.85}  &\underline{72.90}  
&71.25  &\underline{68.26}  &71.27  \\

\midrule
\multicolumn{1}{c|}{\multirow{1}{*}{\textbf{Ours}}} & \textbf{LLaVA-RadZ$_{\textbf{ft}}$} 
& \textbf{73.36} & \textbf{48.59} & \textbf{82.15} 
& \textbf{72.61} & \textbf{27.91} & \textbf{84.64} 
& \textbf{84.36} & \textbf{77.53} & 74.58
& \textbf{86.98} & \textbf{76.18} & \textbf{83.28} 
& \textbf{89.92} & \textbf{79.57} & \textbf{84.38} \\

% \multicolumn{1}{c|}{} & \textbf{LLaVA-RadZ$_{\textbf{full}}$} &  &  &  &  &  &  &  &  &  &  &  &  &  &  &  \\
\bottomrule
\end{tabular}
}
\end{table*}

\begin{table*}[t]
\caption{Comparison of performance with other SOTA methods on four medical datasets for the zero-shot classification task, with AUC, F1, and ACC scores reported. The best results are highlighted in bold and the second-best results are underlined.}
\label{tab:table2}
\resizebox{\textwidth}{!}{
\begin{tabular}{@{}c|ccc|ccc|ccc|ccc@{}}
\toprule
% \multicolumn{1}{c|}{\textbf{Dataset}} 
\multirow{2}{*}{\textbf{Method}}
% & \multicolumn{3}{c|}{\textbf{CheXpert~\citep{irvin2019chexpert}}} 
& \multicolumn{3}{c|}{\textbf{ChestXray-14}} 
& \multicolumn{3}{c|}{\textbf{COVIDx CXR-2}}  
& \multicolumn{3}{c|}{\textbf{RSNA Pneumonia}} 
& \multicolumn{3}{c}{\textbf{SIIM-ACR}}\\ 

\cmidrule(l){2-13} 
% \multicolumn{1}{c|}{\textbf{Method}} 
% & AUC $\uparrow$ & F1 $\uparrow$ & ACC $\uparrow$ 
& AUC $\uparrow$ & F1 $\uparrow$ & ACC $\uparrow$ 
& AUC $\uparrow$ & F1 $\uparrow$ & ACC $\uparrow$ 
& AUC $\uparrow$ & F1 $\uparrow$ & ACC $\uparrow$ 
& AUC $\uparrow$ & F1 $\uparrow$ & ACC $\uparrow$\\
\midrule
\multicolumn{1}{c|}{ConVIRT~\citep{zhang2022contrastive} }
% & 52.10 & 35.61 & 57.43 
& 53.15 & 12.38 & 57.88 
& 62.78 & 71.23 & 63.84 
& 79.21 & 55.67 & 75.08 
& 64.25 & 42.87 & 53.42 \\

\multicolumn{1}{c|}{GLoRIA~\citep{huang2021gloria} }
% & 54.84 & 37.86 & 60.70 
& 55.92 & 14.20 & 59.47 
& 64.52 & 70.78 & 60.21 
& 70.37 & 48.19 & 70.54 
& 54.71 & 40.39 & 47.15 \\

\multicolumn{1}{c|}{BioViL~\citep{boecking2022making} }
% & 60.01 & 42.10 & 66.13 
& 57.82 & 15.64 & 61.33 
& 61.40 & 70.92 & 58.20 
& 84.12 & 54.59 & 74.43 
& 70.28 & 46.45 & 68.22 \\

\multicolumn{1}{c|}{CheXzero~\citep{tiu2022expert} }
% & 87.90 & 61.90 & 81.17 
& 66.99 & 21.99 & 65.38
& 73.13 & 76.13 & 71.45 
& 85.13 & 61.49 & 78.34
& 84.60 & 65.97 & 77.34 \\

\multicolumn{1}{c|}{MedKLIP~\citep{wu2023medklip} } 
% & \underline{87.97} & \underline{63.67} & \underline{84.32} 
& 72.33 & 24.18 & 79.40 
& 76.28 & 76.54 & 71.96 
& 86.57 & 63.28 & 79.97
& 89.79 & 72.73 & 83.99 \\

\multicolumn{1}{c|}{MAVL~\citep{phan2024decomposing} } 
% & \textbf{90.13} & \textbf{65.47} & \textbf{86.44} 
& \textbf{73.50} & \underline{26.25} &  \underline{82.77} 
& \underline{83.86} & \textbf{81.73} & \textbf{78.07} 
& \underline{86.91} & \underline{63.41} & \underline{82.42}
& \textbf{92.04} & \underline{77.95} & \textbf{87.14} \\

\midrule
\multicolumn{1}{c|}{\textbf{Ours}}

% & 75.63 & 48.41 & 81.77
& \underline{72.61} & \textbf{27.91} & \textbf{84.64} 
& \textbf{84.36} & \underline{77.53} & \underline{74.58} 
& \textbf{86.98} & \textbf{76.18} & \textbf{83.28} 
& \underline{89.92} & \textbf{79.57} & \underline{84.38} \\

\bottomrule
\end{tabular}
}
\end{table*}

\begin{table*}[t]
\centering
\caption{Comparison of performance with other SOTA methods at different data portions for fine-tuning classification task.
AUC scores are reported.
The best results are highlighted in bold and the second-best results are underlined.}
\label{tab:table3}
\resizebox{0.95\textwidth}{!}{
\begin{tabular}{@{}c|ccc|ccc|ccc}
\toprule
% \multicolumn{1}{c|}{\textbf{Dataset}} 
\multirow{2}{*}{\textbf{Method}}
& \multicolumn{3}{c|}{\textbf{RSNA Pneumonia }} 
& \multicolumn{3}{c|}{\textbf{Pneumothorax}} 
& \multicolumn{3}{c}{\textbf{COVIDx CXR-2 }} \\
% & \multicolumn{3}{c}{\textbf{ChestXray-14~\citep{wang2017chestx}}} \\ 

\cmidrule(l){2-10} 
% \multicolumn{1}{c|}{\textbf{Data portion}} 
& 1\% & 10\% & 100\% 
& 1\% & 10\% & 100\% 
& 1\% & 10\% & 100\%  \\
% & 1\% & 10\% & 100\%  \\
\midrule
\multicolumn{1}{c|}{Scratch}
& 68.94 & 83.31 & 87.12 
& 53.11 & 76.18 & 87.48 
& 85.11 & 93.65 & 98.86 \\
% & 45.88 & 56.27 & 67.03 \\

\multicolumn{1}{c|}{ConVIRT~\citep{zhang2022contrastive} }
& 78.86 & 85.42 & 87.64 
& 72.39 & 80.41 & 91.67 
& 90.30 & 97.74 & 99.70 \\
% & 57.23 & 72.53 & 79.13 \\

\multicolumn{1}{c|}{GLoRIA~\citep{huang2021gloria} }
& 79.13 & 85.59 & 87.83 
& 75.85 & 86.20 & 91.89 
& 92.74 & 97.18 & 99.54 \\
% & 58.94 & 72.87 & 79.92 \\

\multicolumn{1}{c|}{BioViL~\citep{boecking2022making} }
& 80.27 & 86.04 & 88.29 
& 70.29 & 79.45 & 88.05
& 92.39 & 98.39 & 99.68 \\
% & 60.83 & 72.94 & 80.16 \\

\multicolumn{1}{c|}{MedKLIP~\citep{wu2023medklip} } 
& 82.11 & 87.14 & 88.58 
& 85.24 & 89.91 & 93.02 
& 95.58 & 98.77 & 99.77 \\
% & 62.09 & \underline{74.02} & \underline{80.90} \\

\multicolumn{1}{c|}{MAVL~\citep{phan2024decomposing} } 
& \underline{86.09} & \underline{87.90} & \underline{88.94} 
& \textbf{91.53} & \textbf{93.00} & \underline{94.48} 
& \underline{97.18} & \underline{99.15} & \underline{99.90} \\
% & \textbf{68.65} & \textbf{80.50} & \textbf{86.22} \\

\midrule
\multicolumn{1}{c|}{\textbf{Ours}}
& \textbf{88.23} & \textbf{88.57} & \textbf{89.49} 
& \underline{88.42} & \underline{89.96} & \textbf{94.50} 
& \textbf{98.32} & \textbf{99.80} & \textbf{99.96} \\
% & \underline{68.59} & 70.44 & 73.81 \\

\bottomrule
\end{tabular}
}
\end{table*}
\section{Experiments}
\label{sec:experiments}
In this section, we first provide an overview of the dataset employed in our experiments, including those used for pre-training and the various downstream tasks.
Subsequently, we outline the implementation details and describe the baselines considered for comparison.

\subsection{Dataset}
In our experiments, we pre-trained the model using the MIMIC-CXR dataset~\citep{johnson2019mimic}. For downstream tasks, we primarily evaluated the model’s performance in medical disease classification using multiple benchmark datasets, including ChestX-ray14~\citep{wang2017chestx}, RSNA Pneumonia~\citep{shih2019augmenting}, SIIM-ACR Pneumothorax~\citep{siim2019}, CheXpert~\citep{irvin2019chexpert}, and COVIDx CXR-2~\citep{pavlova2022covid}. Detailed information on these datasets can be found in the supplementary material.

\subsection{Evaluation Metrics} 
For the zero-shot classification task, we employ standard classification evaluation metrics, including Accuracy, AUC score, and F1 score.
The macro-average metrics are reported for all diseases present in the target dataset.

% \subsection{Implementation Details}
% Unless otherwise specified, we use LLaVA-Med~\citep{li2024llava} as the foundational MLLM $\mathcal{F}$.
% % To facilitate segmentation tasks, we integrate a ResUNet as the visual segmentation head.
% % Our training is conducted on two NVIDIA A100 GPUs, each with 80 GB of memory.
% We employ the LoRA strategy for parameter-efficient fine-tuning, with training managed via the DeepSpeed engine.

% For optimization, we utilize the AdamW optimizer with a learning rate of 2e-5 and no weight decay.
% A cosine learning rate decay schedule is applied, with $3\%$ of the total training steps allocated for warm-up.
% The number of special tokens for images $\mathit{<{ImgCls}>}$ is set to 4, while the number of special tokens for text $\mathit{<{TxtCls}>}$ is set to 8.
% The temperature hyperparameter $\tau$ is configured as $0.05$, and the loss weight coefficient $\lambda$ is set to $0.5$.
% Furthermore, the batch size per GPU is set to $64$.

\subsection{Zero-shot evaluation}
As shown in Tab.~\ref{tab:table2}, we compare the performance of established methods in the field on the zero-shot classification task for radiological diseases, evaluated on four officially released test datasets. Our findings demonstrate that, compared to conventional CLIP-style models such as ConVIRT~\citep{zhang2022contrastive}, GLoRIA~\citep{huang2021gloria}, BioViL~\citep{boecking2022making}, and CheXzero~\citep{tiu2022expert}, our approach exhibits significant advantages. Even when compared to state-of-the-art models incorporating external models or domain-specific expert knowledge, our method remains highly competitive. Specifically, on the multi-class dataset ChestXray-14, our model surpasses the supervised learning method MAVL~\citep{phan2024decomposing} by 1.87\% in accuracy. Moreover, on the RSNA Pneumonia dataset, we achieve a 12.77\% improvement in F1 score. These results indicate that multimodal large language models (MLLMs) possess strong feature extraction capabilities, further underscoring their immense potential in medical disease classification tasks.

% 如表2所示，我们比较了当前领域成熟的方法在放射学疾病零样本分类任务上的性能表现，评估数据来自4个官方公布的测试集。我们发现和领域中传统clip风格模型（如ConVIRT，GLoRIA，BioViL,CheXzero ）相比，我们的模型展现出了显著的优势。即使和引入了外部模型或领域专家知识的模型相比，我们的方法同样也有着较强的竞争力，在多分类数据集ChestXray-14上，我们模型的准确率超过了监督学习方法MAVL1.8%，而在RSNA Pneumonia数据集上，我们将F1得分提升了12.43%，这一结果表明MLLMs具有较强的特征捕捉能力，进一步展示了其在医学疾病分类任务上的巨大潜力。

\begin{figure*}[t]
    \center{\includegraphics[width=0.9\textwidth]{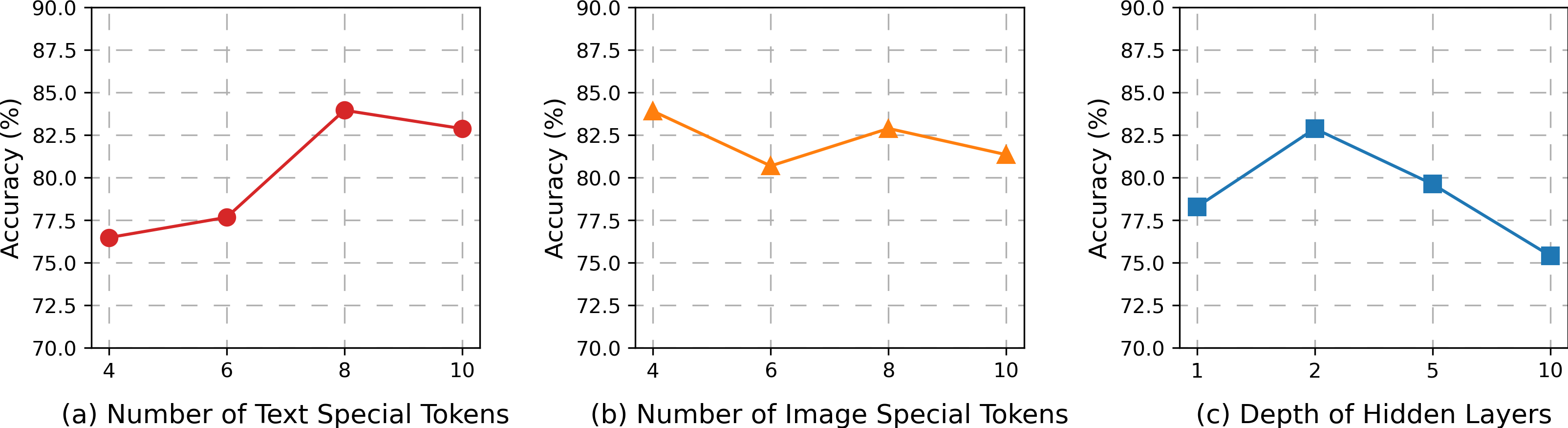}}
    \caption{
        Effect of Special Token Numbers and Hidden Layer Depth on ChestXray-14 Classification.
    }
    \label{fig4}
    % \vspace{-8pt}
\end{figure*}

\subsection{Fine-tuning evaluation}
Consistent with previous studies~\citep{phan2024decomposing,wu2023medklip}, we fine-tune the model on downstream datasets using $1\%$, $10\%$, and $100\%$ of the available data and further evaluate its performance.
Tab.~\ref{tab:table3} presents the fine-tuning results across three datasets, demonstrating that our model consistently maintains a competitive advantage.
Notably, when fine-tuned with only $1\%$ data, our proposed LLaVA-RadZ outperforms the MAVL~\citep{phan2024decomposing} model by $2.14\%$ on the RSNA Pneumonia and by $1.14\%$ on COVIDx.
Even when fine-tuned with $100\%$ data, our model continues to deliver performance improvements.
This enhancement is likely attributed to our decoder-side alignment training strategy, which effectively captures global modality information and leverages the interaction between global and local features to achieve fine-grained cross-modal alignment, further strengthening the model’s disease recognition capability.

% 我们与先前的研究【】保持一致，让模型在下游数据集上分别采用1\%,10\%,100\%的数据量进行微调，并进一步评估模型性能。表3展示了在3个数据集上进行微调的结果，可以发现我们的模型仍然保持着一定的优势。值得注意的是，在使用1\%数据的前提下，我们提出的LLaVA-RadZ在RSNA Pneumonia数据集上比MAVL模型提高了2.03\%,而在COVIDx数据集上也提升了1.16\%，即使使用100\%的数据进行微调，我们的模型依然可以带来一定的提升，这可能是因为我们解码端对齐训练策略能够较好的捕捉到模态的全局信息，并充分利用全局和局部特征之间的交互实现不同模态间的细粒度对齐，进一步增强了模型疾病识别能力。

\begin{table*}[t]
\begin{minipage}{0.46\textwidth} % 左栏
    \centering
    \resizebox{0.96\linewidth}{!}{
    \begin{threeparttable}
    \caption{
        % Impact of Domain Knowledge Anchor Module (DKAM) on the performance of ChestXray-14.
        Ablation study of DKAM on ChestXray-14. 
        ${\textit{D}}_{1}$ represents a semantic vector library of $75$ medical entities, and ${\textit{D}}_{2}$ represents a semantic vector library of $14$ disease categories.
    }
    \label{tab:ablation1}
    \begin{tabular}{c|ccc|ccc}
    \toprule
    \textbf{$\#$} & \textit{DKAM} & ${\textit{D}}_{1}$ & ${\textit{D}}_{2}$ & AUC $\uparrow$ & F1 $\uparrow$  & ACC $\uparrow$ \\ 
    \midrule
    a & &  &  &69.31  &27.30  &82.32 \\
    b & $\checkmark$ & $\checkmark$ &  &68.67 &25.73 &81.84 \\
    c & $\checkmark$ &  & $\checkmark$ &\textbf{72.61} &\textbf{27.91} &\textbf{84.64} \\
    \bottomrule
    \end{tabular}
    \end{threeparttable}
    }
\end{minipage}%
\hfill
\begin{minipage}{0.50\textwidth} % 右栏
    \centering
    \resizebox{0.99\linewidth}{!}{
    \begin{threeparttable}
    \caption{
        % Evaluating the impact of various feature representations on the ChestXray-14.
        Ablation study of feature representations on ChestXray-14.
        % dataset, including global features, local features, and the incorporation of prompts.
    }
    \label{tab:ablation2}
    \begin{tabular}{c|ccc|ccc}
    \toprule
    \textbf{$\#$} & \textit{Global} & \textit{Local} & \textit{Prompt}  & AUC $\uparrow$ & F1$\uparrow$ & ACC$\uparrow$ \\ 
    \midrule
    a & &$\checkmark$  &  &67.14  &25.11  &77.82 \\
    b &  &$\checkmark$ & $\checkmark$ &68.29  &26.42  &78.63  \\
    c & $\checkmark$ &   &$\checkmark$  &70.13  &26.22  &82.50 \\
    d & $\checkmark$ & $\checkmark$  & $\checkmark$ &\textbf{72.61} &\textbf{27.91} &\textbf{84.64} \\
    \bottomrule
    \end{tabular}
    \end{threeparttable}
    }
\end{minipage}
\end{table*}

\subsection{Ablation Study}

\noindent\textbf{Ablation Study of DKAM.} 
To validate the effectiveness of our proposed Domain Knowledge Anchor Module (DKAM), we conducted an ablation study on the ChestXray-14 dataset. With DKAM incorporated, we further investigated the impact of different category semantic vector repositories on the model’s fine-grained alignment capability. Consistent with the previous MedKLIP study, we selected 75 primary medical entities from the MIMIC-CXR dataset. However, unlike MedKLIP, we leveraged the model’s intrinsic domain knowledge to construct a category semantic vector repository, denoted as ${D}_{1}$. Additionally, we built a disease-specific semantic vector repository for the 14 medical disease categories present in the MIMIC-CXR training dataset, denoted as ${D}_{2}$.

As shown in Tab.~\ref{tab:ablation1} (a vs. c), the introduction of DKAM significantly enhances model performance. Using disease category semantics as an intermediary facilitates more precise alignment between medical images and textual descriptions at the category level. Further comparisons in Tab.~\ref{tab:ablation1} (b vs. c) demonstrate that, compared to a larger repository of medical entities, a semantic vector repository focusing on primary disease categories provides stronger guidance for image-text alignment. Moreover, additional medical entities in ${D}_{1}$, such as tip, tube, PICC, and device, may introduce noise and negatively impact alignment at the disease category level. This adverse effect is corroborated by the comparative results in Tab.~\ref{tab:ablation1} (a vs. b).

% 为了验证我们所提出的 Domain Knowledge Anchor Module DKAM模块的有效性，我们在ChestXray-14数据集上进行了消融实验。在使用DKAM的前提下，我们对比了不同类别语义向量库对模型细粒度对齐能力的影响。跟先前Medklip研究一致，我们利用模型自身领域知识为MIMIC-CXR数据集中主要的75个医学实体构建了类别语义向量库，采用D1表示。同样，我们针对训练数据集MIMIC-CXR中涉及的14种医学疾病类别也构建了疾病描述向量库，采用D2表示。
% 表4（a vs. c）的对比结果表明，引入DKAM模块进一步提升了模型的性能。将疾病类别语义作为桥梁，能够较好的引导医学图像和文本实现类别程度上的细粒度对齐，并让模型自身的领域知识得到充分利用。表4（b vs. c）的对比结果表明，相比于数量更多的医学实体而言，主要医学疾病类别语义能够更好的引导图像与文本之间的对齐。D1中额外的医学实体如tip，tube，picc，device等可能会对模型在疾病类别程度上的对齐造成一定的负面效应，这从表4中（a vs. b）的对比结果可以看出。

\noindent\textbf{Ablation Study of Special Tokens.} 
As shown in Tab.~\ref{tab:table1}, we have demonstrated the effectiveness of the Decoding-Side Feature Alignment Training (DFAT) strategy. To further investigate the design of the critical special tokens integral to this approach, we conducted an in-depth analysis on the ChestXray-14 dataset. As illustrated in Fig.~\ref{fig4}, we observed that the number of text and image tokens significantly influences model performance, with both an excessive and an insufficient count potentially resulting in a loss of modal information. Moreover, our study indicates that the optimal global features are not stored in the final hidden layer but rather in the penultimate layer, which may be attributed to the loss of fine-grained information due to deeper feature aggregation, thereby affecting overall performance.

% 表1中已经论证了我们解码端训练对齐策略Decoding-Side Feature Alignment Training (DFAT)的有效性。对于该策略中至关重要的特殊标记的设计，我们在ChestXray-14数据集上进一步进行了分析。如图??所示，我们发现不同数量的文本和图像标记对模型性能影响较为显著，过多或过少的标记数都可能造成一定程度上的模态信息损失。此外，我们发现并不是最后一层隐藏层保存了最佳的全局特征，倒数第二层的特征反而更好，这可能是因为更深层次的特征聚合造成了细粒度信息的丢失，导致性能反而变差。

\noindent\textbf{Ablation Study of Features.} 
During the process of cross-modal alignment, we conducted a detailed analysis of the impact of global and local features on model performance, and further investigated the effectiveness of using prompts, as shown in Tab.~\ref{tab:ablation2}. The experimental results indicate that utilizing only local features yields the poorest performance, while relying solely on global features provides a certain advantage over local features. This may be attributed to the fact that the specially designed tokens for each modality can more precisely capture the global information of the corresponding modality. Moreover, the combination of global and local features achieves the best performance. Additionally, the incorporation of prompts further enhances the model's ability to capture feature information.

% 在不同模态对齐过程中，我们分析了不同类型的特征对模型性能的影响，包括全局特征，局部特征，此外，还对prompt的使用进行了分析，如表5所示。实验结果表明，仅使用局部特征的效果最差，使用全局特征的性能一定程度上优于局部特征，这可能是由于不同模态所单独设计的特殊标记很好的捕捉到了对应模态的全局特征，而同时结合全局特征和局部特征达到了最佳性能。此外，prompt的使用也一定程度上加强了模型对特征信息的捕捉。

\section{Conclusion}
\label{sec:conclusion}
% In this paper, we propose a simple yet effective framework, LLaVA-RadZ, for zero-shot medical disease recognition.
This paper proposes a simple yet effective framework, LLaVA-RadZ, for zero-shot medical disease recognition.
%
% First, we introduce an end-to-end training strategy, Decoding-Side Feature Alignment Training (DFAT), which leverages the characteristics of the MLLM decoder architecture and utilizes specially designed modality-specific tokens to effectively store modality-related information.
First, we introduce an end-to-end decoding-side feature alignment training strategy to leverage the characteristics of the MLLM architecture and effectively store modality-related information.
Additionally, we employ cross-modal contrastive learning to optimize feature alignment across modalities, enhancing the model's cross-modal understanding capabilities.
%
% Furthermore, we propose a Domain Knowledge Anchoring Module (DKAM), which constructs a category-specific semantic vector repository to facilitate category-level alignment between medical images and textual descriptions.
Furthermore, we propose a domain knowledge anchoring Module to facilitate category-level alignment between medical images and textual descriptions.
%
% Experimental results demonstrate that LLaVA-RadZ achieves outstanding performance across multiple publicly available benchmark datasets, highlighting the significant potential of MLLMs in tackling zero-shot radiological disease recognition tasks.
Experimental results demonstrate that LLaVA-RadZ achieves outstanding performance across multiple benchmarks, highlighting the significant potential of MLLMs in tackling zero-shot radiological disease recognition tasks.

% 在本文中，我们提出了一种简单而高效的框架——LLaVA-RadZ，以实现零样本医学疾病识别。首先，我们设计了一种端到端的训练策略 Decoding-Side Feature Alignment Training (DFAT)，该策略充分利用 MLLM 解码器架构的特性，并结合专为不同模态设计的特殊标记，以有效存储对应模态的信息。同时，我们通过跨模态对比学习优化模态间的特征对齐，从而增强模型的跨模态理解能力。此外，我们提出了一种 Domain Knowledge Anchoring Module (DKAM)，通过构建类别语义向量库进一步引导医学图像与文本在类别级别上的精准对齐。实验结果表明，LLaVA-RadZ 在多个公开测试集上均取得了优异的性能，充分验证了 MLLM 在零样本放射学疾病识别任务中的强大潜力。
% \clearpage

% \section*{Ethics Statement}
% This study follows the ICLR Code of Ethics. All experiments and data usage comply with relevant laws, regulations, and ethical requirements. The data used are from publicly available datasets or obtained with proper authorization, and have been appropriately preprocessed to ensure privacy and security. This work aims to advance scientific research and is not intended for any harmful or inappropriate applications. The authors declare no conflict of interest.

% \section*{Reproducibility Statement}
% We provide detailed descriptions of our methods, datasets, and experimental settings in the main text and appendix. All source code and data processing scripts will be made publicly available upon publication to facilitate reproducibility. Additional implementation details are available in the supplementary materials.
% \newpage
{
    \small
    \bibliography{iclr2026_conference}
    \bibliographystyle{iclr2026_conference}
}

\clearpage

\appendix

\section{Related Work}
\label{sec:related_work}

\textbf{Multi-modal Large Language Models.}
Inspired by the exceptional reasoning capabilities of large language models (LLMs), researchers are actively exploring ways to extend these abilities to the visual domain, driving advancements in multimodal LLMs. With the release of GPT-4 (Vision)~\citep{achiam2023gpt} and Gemini~\citep{team2023gemini}, these models have demonstrated remarkable multimodal understanding and generation capabilities, further fueling research in this field.

To bridge the gap between vision encoders and LLMs, BLIP-2~\citep{li2023blip} introduces a Q-Former that transforms image features into a format compatible with LLMs, enabling seamless integration with text embeddings.
LLaVA~\citep{liu2023visual} and MiniGPT-4~\citep{zhu2023minigpt} further enhance generalization and task performance by leveraging large-scale multimodal pretraining, followed by instruction tuning for specific applications.
In the medical domain, LLMs have shown immense potential for advancing research and practical applications. Med-Flamingo~\citep{moor2023med} extends Flamingo to the medical field by pretraining on multimodal knowledge sources spanning multiple medical disciplines.
LLaVA-Med~\citep{li2024llava} refines image-text pairs from PMC-15M~\citep{zhang2023large} and trains a biomedical-specialized MLLM using a limited dataset, building upon the pre-trained parameters of LLaVA.
Similarly, Med-PaLM~\citep{singhal2023large} fine-tunes PaLM~\citep{chowdhery2023palm} using domain-specific medical instructions, demonstrating strong performance under human evaluation frameworks.
Other notable models, such as Chat-Doctor~\citep{li2023chatdoctor} and Med-Alpaca~\citep{han2023medalpaca}, have been tailored for medical question-answering and dialogue applications.

Despite the significant progress of MLLMs, several challenges remain~\citep{mckinzie2024mm1,tong2024eyes, zhang2024visually,he2025analyzing}.
Recent studies~\citep{zhang2024visually,he2025analyzing} highlight the suboptimal performance of MLLMs in image classification, particularly in fine-grained category recognition.
We find that this issue is especially pronounced in the medical domain, where precise classification is crucial for medical applications.
To address these shortcomings, we are refining traditional MLLM training paradigms to enhance classification performance and improve fine-grained category comprehension.

% 受大型语言模型（LLMs）卓越推理能力的启发，研究人员正积极探索如何将这些能力扩展到视觉领域，推动多模态 LLMs（MM-LLMs）的发展。随着 GPT-4 (Vision)（OpenAI，2023）和 Gemini（Team 等，2023）的发布，这些模型在多模态理解和生成方面展现出卓越能力，进一步推动了该领域的研究进程。

% 为了弥合视觉编码器与 LLM 之间的差距，BLIP-2 [24] 引入了 Q-Former，将图像特征转换为 LLM 可处理的格式，从而实现与文本嵌入的无缝融合。LLaVA [29] 和 MiniGPT-4 [63] 通过大规模多模态预训练，并结合指令微调，显著提升了泛化能力和任务表现。在医学领域，LLMs 也展现出巨大的研究和应用潜力。Med-Flamingo [84] 通过在涵盖多个医学学科的多模态知识源上进行预训练，将 Flamingo 扩展至医学领域。LLaVA-Med [71] 基于 LLaVA 的预训练参数，在经过筛选的 PMC-15M [126] 子集上进行微调，以有限的数据训练出专注于生物医学的多模态 LLM。此外，Med-PaLM（Singhal 等，2023）通过针对医学领域的指令微调 PaLM（Chowdhery 等，2023），在人工评估框架下展现出卓越的性能。同样，Chat-Doctor（Yunxiang 等，2023）和 Med-Alpaca（Han、Adams 等，2023）等模型专为医学问答和对话应用进行了优化。

% 尽管 MLLMs 取得了显著进展，但仍面临诸多挑战。研究表明，这些模型在图像分类任务上表现不佳，尤其是在细粒度类别识别方面存在局限性。这一问题在医学领域尤为突出，精确分类对医学应用至关重要。为了解决这些不足，我们正在改进传统的 MLLM 训练范式，以增强其分类能力，并提升对细粒度类别的理解。

\textbf{Prompt Engineering.}
Prompting enhances the ability of pre-trained large language models (LLMs) to understand tasks by incorporating language instructions into the input text~\citep{mondal2024kam,shao2024visual,liu2024era,li2024focus}.
Recently, prompt-based techniques have also been applied to vision-language models to improve performance.
In medical vision-language models (VLMs), GloRIA~\citep{huang2021gloria} generates a set of textual prompts to describe potential subtypes, severity levels, and anatomical locations for each disease category.
MedKLIP~\citep{wu2023medklip} enhances model performance by retrieving descriptions of medical entities from the UMLS knowledge base~\citep{bodenreider2004unified}.
CARZero~\citep{lai2024carzero} introduces a prompt-alignment strategy based on LLMs, integrating prompt templates into the training dataset to ensure alignment during both training and inference.
MAVL~\citep{phan2024decomposing} uses visual descriptions of pathological features to guide the model in effectively detecting diseases in medical images.

Although these approaches have successfully improved model performance through prompt-based strategies, they all rely on external models or expert knowledge, without fully leveraging the model’s intrinsic understanding capabilities.
Fortunately, recent research on LLaVA-Med~\citep{li2024llava} has demonstrated remarkable domain-specific conversational abilities, proving that it possesses a certain level of medical knowledge.
Building upon LLaVA-Med~\citep{li2024llava}, we further explore the feasibility of utilizing the model’s inherent comprehension to enhance zero-shot medical classification performance.
% 提示（Prompting）通过将语言指令添加到输入文本中，使预训练的语言模型（LLM）能够理解任务。最近，提示技术也被应用于视觉-语言模型以提升模型性能。在医学VLM中，GloRIA 生成一组文本提示来描述每种疾病类别中可能的亚型、严重程度和位置。MedKLIP通过从UMLS知识库中获取每个医学实体的描述来增强模型性能。CARZero提出了一种基于LLM的提示对齐策略，将提示模板集成到训练数据集中实现训练和推理阶段的提示对齐。MAVL提出利用病理的视觉描述来指导模型有效地检测图像中的疾病。尽管上述方法提出的提示策略都一定程度上提升了模型性能，但都引入了外部模型或专家的领域知识，并没有充分利用模型自身的理解能力。幸运的是，最近的研究LLaVA-Med展现了出色的领域知识对话能力，证明其具备了一定的医学领域知识。我们在LLaVA-Med的研究基础上进一步探索了利用模型自身理解能力来提升医学零样本分类性能的可行性。

% \clearpage

% \maketitle
% \clearpage

% \onecolumn
% \clearpage

% \onecolumn
% 疾病描述提示模版
% \appendix
% \section{Appendix}
% \subsection{Dataset Details}
\section{Dataset Details}
\label{appendix:dataset}
\textbf{MIMIC-CXR v2~\citep{johnson2019mimic}.}
In our experiments, we pre-trained the model using the MIMIC-CXR, a publicly available collection of chest radiographs paired with corresponding radiology text reports.
The MIMIC-CXR dataset comprises $377,110$ images corresponding to 227,835 radiographic studies from 65,379 patients.
Since all downstream tasks utilize frontal-view images, we exclude all lateral-view images from the dataset.
Moreover, we selectively retain only the findings and impressions sections from these reports.

\noindent\textbf{ChestX-ray14~\citep{wang2017chestx}.}
ChestX-ray14 consists of $112,120$ frontal-view chest X-ray images from 30,805 unique patients, collected between 1992 and 2015.
The official test set, comprising $22,433$ images, has been meticulously annotated by board-certified radiologists.
For evaluation purposes, we restrict our testing to the official test set.

\noindent\textbf{RSNA Pneumonia~\citep{shih2019augmenting}.}
RSNA Pneumonia includes over $260,000$ frontal-view chest X-rays with annotated pneumonia masks, collected by the Radiological Society of North America (RSNA).
This dataset supports both pneumonia segmentation and classification tasks~\citep{wu2023medklip,phan2024decomposing}.
We partition the dataset into training, validation, and test sets with a ratio of $0.6 / 0.2 / 0.2$, respectively.

\noindent\textbf{SIIM-ACR Pneumothorax~\citep{siim2019}.} SIIM-ACR Pneumothorax contains 12,954 chest X-ray images, along with image-level pneumothorax annotations and pixel-level segmentation masks where pneumothorax is present.
Like the RSNA Pneumonia dataset, it can be used for both classification and segmentation tasks.
We divide the dataset into training, validation, and test sets with a ratio of $0.6 / 0.2 / 0.2$.

\noindent\textbf{CheXpert~\citep{irvin2019chexpert}.}
CheXpert contains 224,316 chest X-ray images from 65,240 patients, collected by Stanford Hospital.
The official test set includes images from $500$ patients, annotated through consensus by five board-certified radiologists. We evaluated all disease categories in this test dataset.

\noindent\textbf{COVIDx CXR-2~\citep{pavlova2022covid} and COVID Rural~\citep{desai2020chest}.}
The COVIDx CXR-2 and COVID Rural are designed for evaluating COVID-19 diagnosis.
COVIDx CXR-2~\citep{pavlova2022covid} consists of 29,986 images from 16,648 COVID-19 patients, each labeled with a classification tag. The dataset is split into training, validation, and test sets with a ratio of $0.7 / 0.2 / 0.1$, used for evaluating classification performance.
The COVID Rural dataset contains over $200$ chest X-ray images with annotated segmentation masks, used for the COVID-19 segmentation task.
This dataset is partitioned into training, validation, and test sets with a ratio of $0.6 / 0.2 / 0.2$.

% \subsection{Medical Category Semantic Vector Library}
\section{Medical Category Semantic Vector Library}
\label{appendix:Category}
We draw inspiration from the work of MedKLIP~\citep{wu2023medklip} and incorporate 75 frequently occurring medical entities from clinical reports as input to our model. By designing prompts, we stimulate the model’s intrinsic medical knowledge, enabling it to infer the semantic representations of various entity categories. The resulting semantic descriptions of these 75 medical entities are presented in ~\cref{tab:entity_descriptions}.

Furthermore, to achieve a more precise representation of major disease categories, we construct a dedicated disease semantic vector library, which facilitates a more nuanced understanding of disease-related semantics. The generated disease descriptions are detailed in ~\cref{tab:disease_descriptions}.
% 我们参考了Medklip的工作，将报告中经常出现的75种医学实体输入我们的模型，通过设计prompt激发模型自身的医学领域知识挖掘实体类别语义，生成的75种实体类别语义如表7所示。此外，我们针对主要疾病类别单独构建了疾病语义向量库，生成的疾病描述如表6所示。

% \subsection{Implementation Details}
\section{Implementation Details}
Unless otherwise specified, we use LLaVA-Med~\citep{li2024llava} as the foundational MLLM $\mathcal{F}$.
% To facilitate segmentation tasks, we integrate a ResUNet as the visual segmentation head.
% Our training is conducted on two NVIDIA A100 GPUs, each with 80 GB of memory.
We employ the LoRA strategy for parameter-efficient fine-tuning, with training managed via the DeepSpeed engine.

For optimization, we utilize the AdamW optimizer with a learning rate of 2e-5 and no weight decay.
A cosine learning rate decay schedule is applied, with $3\%$ of the total training steps allocated for warm-up.
The number of special tokens for images $\mathit{<{ImgCls}>}$ is set to 4, while the number of special tokens for text $\mathit{<{TxtCls}>}$ is set to 8.
The temperature hyperparameter $\tau$ is configured as $0.05$, and the loss weight coefficient $\lambda$ is set to $0.5$.
Furthermore, the batch size per GPU is set to $64$.

% \subsection{Use of LLMs} 
% \section{Use of LLMs} 
% This paper employed large language models (\ie, ChatGPT, Claude) solely for language editing and polishing purposes, including but not limited to grammar checking, expression optimization, and text refinement. All core research content, including experimental design, data analysis, and conclusion derivation, was carried out independently by the authors. The authors take full responsibility for the entire content of this paper and have thoroughly verified and validated all AI-assisted modifications.

\newpage
\begin{longtable}{@{}p{3cm}p{10cm}@{}}
\caption{Semantic Descriptions of 14 Medical Disease Categories}
\label{tab:disease_descriptions} \\
\toprule
\textbf{Disease} & \textbf{Description} \\
\midrule
\endfirsthead

% 续页使用相同的标题（去掉label避免重复标签警告）
\caption{Semantic Descriptions of 14 Medical Disease Categories} \\
\toprule
\textbf{Disease} & \textbf{Description} \\
\midrule
\endhead

% 页面底部：只显示水平线，不显示"Continued"文字
\midrule
\endfoot

% 表格结束时的底部规则
\bottomrule
\endlastfoot

Fibrosis & Fibrosis refers to excessive deposition of collagen and extracellular matrix during abnormal tissue repair after inflammation or injury, leading to the replacement of normal lung tissue with reticular or band-like high-density shadows, commonly seen in the lower and peripheral lungs. Imaging may show honeycombing and traction bronchiectasis. Clinically, patients often present with progressive dyspnea, dry cough, and reduced exercise tolerance. \\

Edema & Pulmonary edema refers to the abnormal accumulation of fluid in the pulmonary interstitium and alveoli, usually caused by cardiogenic or non-cardiogenic factors. Imaging shows patchy or 'bat-wing' distributed heterogeneous high-density shadows in the middle or entire lung, often accompanied by Kerley lines and cardiac enlargement. Clinically, patients typically experience acute dyspnea, cough, cyanosis, and bilateral lung crackles. \\

Pneumothorax & Pneumothorax refers to the presence of air in the pleural cavity, leading to partial or complete lung collapse. Imaging typically shows a low-density black air space along the pleura, with a clear demarcation from the normal lung tissue, along with lung collapse. In tension pneumothorax, mediastinal shift may occur. Clinically, patients often present with sudden unilateral chest pain, dyspnea, and decreased breath sounds, sometimes accompanied by subcutaneous emphysema. \\
Cardiomegaly & Cardiomegaly refers to the enlargement of the heart due to hypertension, cardiomyopathy, or valvular disease, causing chamber dilation or wall thickening. Imaging shows significant cardiac enlargement with an expanded and smooth contour, often marked by an increased cardiothoracic ratio, potentially accompanied by pulmonary congestion and bronchial congestion. Clinically, patients may experience reduced exercise tolerance, dyspnea, lower limb edema, and arrhythmias. \\
Atelectasis & Atelectasis refers to the collapse of part or all of the lung tissue due to airway obstruction, external thoracic pressure, or intrapulmonary pathology. Imaging shows increased local lung density, volume reduction, bronchial displacement, and visceral pleural traction, commonly affecting the lower lobes. Clinically, patients may exhibit rapid shallow breathing, localized decreased or absent breath sounds, and a history of recent surgery or inadequate airway clearance. \\
Nodule & A lung nodule is a localized lesion with a diameter of less than 3 cm. Imaging typically shows a round or oval localized density, with either well-defined or spiculated edges. Some nodules may contain calcifications or low-density necrotic areas. Clinically, most patients are asymptomatic, but growing or malignant nodules may present with cough and hemoptysis. \\
Emphysema & Emphysema is a chronic obstructive pulmonary disease caused by the permanent destruction of alveolar walls and airspace enlargement. Imaging shows scattered or diffuse low-density areas in both lungs, reduced lung markings, often with bullae or cystic lesions, a flattened diaphragm, and hyperinflated lungs. Clinically, patients typically have a history of chronic cough, sputum production, and progressive dyspnea, often associated with smoking or long-term occupational exposure. \\
No Finding & No finding refers to the absence of radiographic abnormalities detected in the chest X-ray. \\
Mass & A mass refers to an abnormal localized tissue overgrowth. Imaging shows a focal high-density lesion, which may have regular or irregular shapes with spiculated margins, often accompanied by internal necrosis, calcification, or hemorrhage. Surrounding features may include bronchial distortion or lymphadenopathy. Clinically, patients may present with cough, weight loss, or hemoptysis, requiring further pathological examination. \\
Pleural Thickening & Pleural thickening refers to fibrotic or calcified pleural changes due to chronic inflammation, infection, or asbestos exposure. Imaging shows localized or diffuse thickening along the pleural surface, appearing as streaky or patchy high-density shadows, sometimes with nodular changes. Clinically, patients may be asymptomatic, but a history of pleuritis or exposure to harmful substances is often present. \\

Effusion & Pleural effusion refers to the abnormal accumulation of fluid in the pleural cavity, which may be caused by infection, heart failure, malignancy, or other inflammatory diseases. Typically seen in the lower lung fields and posterior chest cavity, imaging shows a homogeneous or layered fluid density with a clear meniscus sign, with CT revealing low-density regions. Severe effusion may cause lung compression or bronchial displacement. Clinically, patients may present with dyspnea, chest pain, and cough, with physical signs of reduced breath sounds, dull percussion, and abnormal auscultation. \\
Infiltration & Infiltration refers to localized or diffuse high-density changes in lung tissue due to inflammation, infection, or malignancy. Imaging typically shows patchy or ill-defined high-density areas, sometimes with a ground-glass appearance or consolidation, occasionally accompanied by air bronchograms or bronchial wall thickening. Clinically, patients may present with cough, fever, dyspnea, and fatigue, often with elevated inflammatory markers. \\
Pneumonia & Pneumonia refers to lung parenchyma inflammation caused by bacteria, viruses, fungi, or other microorganisms, leading to alveolar filling with inflammatory exudates. Imaging shows localized or patchy consolidation with irregular margins, often accompanied by air bronchograms, pleural reaction, and mild pleural effusion. Clinically, patients present with fever, cough, sputum production, chest pain, and fatigue, with elevated white blood cell counts and inflammatory markers. \\
% Hernia & Hernia refers to the abnormal protrusion of an organ or tissue through an anatomical defect or weakened area, commonly seen in hiatal hernias. Imaging shows intra-abdominal organs abnormally positioned in the thoracic cavity, appearing as continuous soft-tissue density or gas-fluid levels, often with diaphragmatic defects or abnormalities. Clinically, patients may present with upper abdominal discomfort, acid reflux, dysphagia, or chest pain, often associated with gastroesophageal reflux disease or digestive issues. \\
Consolidation & Consolidation refers to the complete filling of alveolar spaces with liquid, pus, blood, or cellular material, replacing the normal air content. Imaging shows homogeneous, dense, well-defined opacities, often with air bronchograms and pleural reactions, sometimes with minimal pleural effusion. Clinically, patients often have fever, cough, sputum production, chest pain, and dyspnea, with significantly elevated inflammatory markers. \\

\end{longtable}

\newpage
\centering
\begin{longtable}{@{}p{3cm}p{10cm}@{}}
\caption{Semantic Descriptions of 75 Medical Categories}
\label{tab:entity_descriptions} \\
\toprule
\textbf{Disease} & \textbf{Description} \\
\midrule
\endfirsthead

% 续页使用相同的标题（去掉label避免重复标签警告）
\caption{Semantic Descriptions of 75 Medical Categories} \\
\toprule
\textbf{Disease} & \textbf{Description} \\
\midrule
\endhead

% 页面底部：只显示水平线，不显示"Continued"文字
\midrule
\endfoot

% 表格结束时的底部规则
\bottomrule
\endlastfoot

normal & Indicates that the structure appears within standard parameters without signs of pathology. \\
clear & The imaging reveals no obscuring abnormalities, ensuring clear visualization of the structure. \\
sharp & Boundaries are precisely defined, accentuating the distinct separation between tissues. \\
sharply & The structure is rendered with exceptional clarity, facilitating detailed evaluation. \\
unremarkable & No significant deviations or abnormalities are observed in the examined area. \\
intact & The structure remains whole and undamaged, with no disruption detected. \\
stable & The tissue exhibits consistent appearance over time without progressive changes. \\
free & Presence of extraluminal air in unexpected locations, possibly indicating a perforation. \\
effusion & Accumulation of fluid between the pleural layers, often reflecting an underlying pathology. \\
opacity & An area of increased radiodensity that obscures normal lung markings, suggesting fluid or tissue replacement. \\
pneumothorax & Air present in the pleural space that may lead to partial or complete lung collapse. \\
edema & Diffuse fluid accumulation within lung tissue, frequently associated with cardiac or inflammatory issues. \\
atelectasis & Collapse of lung segments resulting in volume loss and increased local density. \\
tube & A medical tube visible on imaging, such as for drainage or airway management. \\
consolidation & Region where alveolar air is replaced by fluid or cells, producing homogeneous density. \\
process & Denotes an active pathological condition altering the tissue’s normal appearance. \\
abnormality & A generic term for any deviation from normal structure suggestive of disease. \\
enlarge & Indicates that a structure appears larger than typical normal values. \\
tip & The distal or pointed end of a structure or medical device. \\
low & Underinflation of the lungs, often implying a restrictive process. \\
pneumonia & Inflammatory infection of lung parenchyma, typically showing consolidation and air bronchograms. \\
line & A linear structure that may represent a fissure, pleural interface, or artifact. \\
congestion & Increased blood or fluid accumulation in tissues, often indicating impaired circulation. \\
catheter & A slender, flexible tube inserted for drainage or medication delivery, visible in imaging. \\
cardiomegaly & An enlarged cardiac silhouette, frequently associated with chronic heart conditions. \\
fracture & A break or discontinuity in bone structure evident on radiographs. \\
air & Regions of radiolucency indicating the presence of gaseous content. \\
tortuous & Describes a vessel or structure exhibiting excessive curvature or winding. \\
lead & The foremost or guiding portion of a device or anatomical feature. \\
disease & A general term for any pathological process affecting normal tissue function. \\
calcification & Deposition of calcium salts within tissue, appearing as bright foci on imaging. \\
prominence & An area that appears more pronounced than surrounding tissues, suggesting an increase in size or density. \\
device & Any implanted or externally applied apparatus used for diagnostic or therapeutic purposes. \\
engorgement & Excessive filling of vessels or tissues with blood, leading to a swollen appearance. \\
picc & A long, thin catheter introduced via a peripheral vein and advanced into the central circulation for long-term therapy. \\
clip & A small metallic or plastic fastener used during surgery to secure tissues or vessels. \\
elevation & An upward displacement or raised position of an anatomical structure relative to its usual location. \\
expand & Describes a structure that appears dilated or increased in volume. \\
nodule & A small, rounded lesion typically less than 3 cm in diameter that can be benign or malignant. \\
wire & A thin, flexible metallic strand often used in surgical fixation or as part of medical devices. \\
fluid & The presence of liquid within tissues or cavities, altering the normal radiographic appearance. \\
degenerative & Changes in tissue structure resulting from chronic wear, aging, or repeated stress. \\
pacemaker & An implanted device that regulates heart rhythm, visible through its leads and generator. \\
thicken & Describes a structure that appears denser or more layered, possibly due to fibrotic changes. \\
marking & Visible patterns or lines that may represent vascular or connective tissue features. \\

scar & Fibrotic tissue that replaces normal parenchyma following injury, typically seen as an irregular opacity. \\
hyperinflate & Denotes lungs that are over-expanded, often with increased radiolucency and flattened diaphragms. \\
blunt & Loss of sharp definition in anatomical borders, leading to a less distinct appearance. \\
loss & Indicates a reduction or absence of normal tissue volume or density. \\
widen & Suggests that a structure or space is broader than the standard measurement. \\
collapse & A significant reduction or complete loss of volume in lung tissue due to obstruction or injury. \\
density & Reflects the compactness of a tissue, with higher density appearing whiter on radiographs. \\
emphysema & A chronic condition marked by alveolar wall destruction and abnormal enlargement of air spaces. \\
aerate & Indicates that the lung tissue is adequately filled with air, supporting effective gas exchange. \\
mass & A malignant tumor arising from lung tissue, typically presenting as an irregular mass with possible cavitation. \\
crowd & Compaction of airways and vessels, often due to volume loss or infiltrative processes. \\
infiltrate & Diffuse or patchy opacities in the lung that suggest inflammation, infection, or neoplastic involvement. \\
obscure & Describes anatomical structures that are not clearly visualized, often due to overlapping tissues or technical factors. \\
deformity & An abnormal shape or structure resulting from congenital anomalies, trauma, or disease progression. \\
hernia & The protrusion of an organ or tissue through an abnormal opening in the surrounding structure. \\
drainage & The process or presence of fluid removal from a body cavity, often via an inserted tube. \\
distention & Abnormal expansion or swelling of a structure due to accumulation of fluid or gas. \\
shift & Displacement of anatomical structures from their usual positions, indicating mass effect or volume change. \\
stent & A small mesh tube used to maintain the patency of a vessel or duct. \\
pressure & The force exerted per unit area by fluids or tissues, which can influence organ function. \\
lesion & Any abnormal area of tissue that deviates from the standard architecture, potentially indicative of pathology. \\
finding & A generic term for an observed abnormality or noteworthy feature on imaging. \\
borderline & The heart appears at the upper limit of normal size, without clear evidence of enlargement. \\
hardware & Any implanted or externally attached device used for diagnostic, therapeutic, or supportive purposes. \\
dilation & The widening or expansion of a hollow structure, often reflecting increased internal pressure. \\
chf & A clinical syndrome characterized by the heart's reduced pumping ability, leading to systemic fluid accumulation. \\
redistribution & A shift in the normal pattern of blood or air distribution within the lungs, often due to altered hemodynamics. \\
aspiration & Inhalation of foreign material into the airways, potentially leading to inflammatory or infectious complications. \\
rare diseases & Conditions that occur infrequently in the population and often require specialized diagnostic and management approaches. \\
Covid-19 & An infectious disease caused by the SARS-CoV-2 virus, with a broad spectrum of respiratory and systemic manifestations. \\
\end{longtable}

\end{document}